\documentclass[10pt,journal,compsoc]{IEEEtran}
\usepackage{amssymb}
\usepackage{amsmath}
\usepackage{microtype}
\usepackage{pgfplots, pgfplotstable, filecontents}
\usepackage{tikz}
\usetikzlibrary{patterns}

\usepackage{adjustbox}
\usepackage[linesnumbered,ruled]{algorithm2e}
\usepackage{setspace}

\SetCommentSty{mycommfont}
\SetKwRepeat{Do}{do}{while}%

\usepackage{graphicx}
\usepackage{subfig}
\usepackage{multirow}
\usepackage{booktabs}

\usepackage{threeparttable}
\usepackage{booktabs,makecell,siunitx}

\newcommand\copyrighttext{%
	\footnotesize This work has been submitted to the IEEE for possible publication. Copyright may be transferred without notice, after which this version may no longer be accessible.}
\newcommand\copyrightnotice{%
	\begin{tikzpicture}[remember picture,overlay]
		\node[anchor=south,yshift=10pt] at (current page.south) {\fbox{\parbox{\dimexpr\textwidth-\fboxsep-\fboxrule\relax}{\copyrighttext}}};
	\end{tikzpicture}%
}

\ifCLASSOPTIONcompsoc
\usepackage[nocompress]{cite}
\else
\usepackage{cite}
\fi
\graphicspath{{fig/}}

\ifCLASSINFOpdf
\else
\fi
\begin{document}
	%
	\title{Metric Learning as a Service with Covariance Embedding}
	%
	%
	%
	%
	
	\author{Imam~Mustafa~Kamal,
		Hyerim~Bae,
		and~Ling~Liu
		\IEEEcompsocitemizethanks{\IEEEcompsocthanksitem IM. Kamal was with the Institute of Intelligent Logistics Big Data, Pusan National University, 30-Jan-jeon Dong, Geum-Jeong Gu, 609-753, Busan, South Korea, Email: imamkamal@pusan.ac.kr.
			\IEEEcompsocthanksitem H. Bae was with a major in Industrial Data Science \& Engineering, Department of Industrial Engineering, Pusan National University, 30-Jan-jeon Dong, Geum-Jeong Gu, 609-753, Busan, South Korea, Email: hrbae@pusan.ac.kr (Corresponding
			author).
			\IEEEcompsocthanksitem L. Liu was with the College of Computing, Georgia Institute of Technology, 266 Ferst Drive, Atlanta, 30332-0765, GA, USA, Email: lingliu@cc.gatech.edu.
		}}

	\IEEEtitleabstractindextext{%
		\begin{abstract}
			With the emergence of deep learning, metric learning has gained significant popularity in numerous machine learning tasks dealing with complex and large-scale datasets, such as information retrieval, object recognition and recommendation systems. Metric learning aims to maximize and minimize inter- and intra-class similarities. However, existing models mainly rely on distance measures to obtain a separable embedding space and implicitly maximize the intra-class similarity while neglecting the inter-class relationship. We argue that to enable metric learning as a service for high-performance deep learning applications, we should also wisely deal with inter-class relationships to obtain a more advanced and meaningful embedding space representation. In this paper, a novel metric learning is presented as a service methodology that incorporates covariance to signify the direction of the linear relationship between data points in an embedding space. Unlike conventional metric learning, our covariance-embedding-enhanced approach enables metric learning as a service to be more expressive for computing similar or dissimilar measures and can capture positive, negative, or neutral relationships. Extensive experiments conducted using various benchmark datasets, including natural, biomedical, and facial images, demonstrate that the proposed model as a service with covariance-embedding optimizations can obtain higher-quality, more separable, and more expressive embedding representations than existing models.
		\end{abstract}
		
		}

	\maketitle
	\copyrightnotice
	
	\IEEEdisplaynontitleabstractindextext

	%
	\IEEEpeerreviewmaketitle

	\IEEEraisesectionheading{\section{Introduction}
		\label{sec:introduction}}


	%
	%
	%
	%
	
	\IEEEPARstart{D}{eep} learning has penetrated many businesses, science, and engineering domains with a remarkable generalization performance, including self-driving cars, smart manufacturing, smart cities, healthcare diagnostics, spam detection, and cloud resource management. The notion of a distance or similarity measure plays a crucial role in deep-learning algorithms. Conventionally, standard distance metrics, such as Euclidean, cosine, and Mahalanobis, are applied by incorporating \textit{a priori} knowledge of the domain. However, it is often challenging to devise metrics that are pertinent to particular data and tasks of interest. 
	
	Metric learning aims to automatically construct task-specific distance or similarity metrics using weakly supervised data in a machine-learning fashion. The outcome can be used to perform various tasks, such as $k$-nearest neighbor ($k$-NN), clustering, and information retrieval. Moreover, metric learning can offer a natural solution to various machine learning problems because it has numerous benefits, including robustness to noisy data, a high generalization to unseen categories, the capability of a dimensionality-reduction model, reliability as a feature extraction model (for fine-tuning classification), and the ability to operate with small sample datasets \cite{7407637}. Accordingly, research into metric learning has received increasing attention over the past few years and has been applied to many different fields, such as information retrieval \cite{LIU2021103067}, recommender systems \cite{WU2020308}, social media mining \cite{8049264}, face recognition \cite{6926840}, and speech recognition \cite{8998226}, to name a few. Siamese, Triplet, and N-pair networks are prominent metric learning frameworks. A Siamese network is trained using two tandem networks with two different input data to compute their similarity. A triplet network is trained using three networks to define the similarity between three images, two of which will be similar (anchor and positive samples), and the third will be unrelated (a negative example). The N-pair network is trained using the cosine similarity to calculate the pairwise distance of N samples, where N $>$ 3. Each of these networks is able to project high-dimensional data into a low-dimensional embedding space while preserving data separability. The samples of different categories must be dissimilar (far from each other); otherwise, the samples from the same categories must have a similar latent representation. Defining the closeness and remoteness among data or samples in complex data, such as natural images, is a non-trivial task. Accordingly, the similarity metric is critical for obtaining a meaningful embedding space representation.
	
	When similarity measures are not given \textit{a priori}, although a generic function, such as Euclidean distance, can be adopted, doing so frequently produces unsatisfactory results \cite{NIPS2012_59b90e10}. If two data vectors have no attribute values in common, they may have a smaller distance than the other pair of data vectors containing the same attribute values. In addition, as one of the main drawbacks of the cosine similarity, the magnitude of the vectors is not considered, and only their direction is considered \cite{NEURIPS2020_d9812f75}. In practice, this means that the differences in values are not fully considered. To the best of our knowledge, existing metric learning models only explicitly maximize intra-class similarity and implicitly neglect the interclass relationship. Thus, the latent space representation is possibly less meaningful because it cannot capture inter-class connections. Moreover, existing metric learning models such as Siamese, Triplet, and N-pair networks often suffer from a low performance under a particular condition. The Siamese network results in fewer semantic outcomes because it neglects the inter-class relationship. The triplet network can diverge when the positive and negative samples have similar features. However, the N-pair network requires a large number of batches to obtain an appropriate result that is computationally expensive. 
	
	In this paper, we propose a novel metric learning method called CovNet with three original contributions. First, CovNet employs covariance to determine the relationship between two pairs of inputs. The covariance signifies the direction of the linear relationship between the two vectors. By direction, we are referring to whether the variables are directly or inversely proportional to each other. Increasing the value of one variable might have a positive or negative impact on the value of the other variable. In addition, unlike the cosine similarity, the covariance subtracts the means before taking the dot product, making it invariant to shifts. Second, we provide three types of mapping functions to embody the covariance embedding: inner-class mapping (IM), inner-class with semi-inter-class mapping (ISIM), and inner- and inter-class mapping (IIM). Finally, we conducted extensive experiments on several benchmark datasets and demonstrated that CovNet outperforms existing deep metric learning models, such as Siamese, Triplet, and N-pair networks. The covariance metric is more expressive than the Euclidean and cosine similarity metrics because it can capture three possible relationships between two variables: a positive, neutral, or negative correlation. Compared with existing metric learning models, as it can obtain accurate inner and inter-class relationships, we show that CovNet is more semantic and expressive.
	
	The remainder of this paper is organized as follows. Section \ref{sec:related_work} reviews the literature on metric learning models. Section \ref{sec:method} presents the proposed method. An extensive experiment along with a discussion is presented in Section \ref{sec:experiment}, and finally, we outline some conclusions and future directions for this research in Section \ref{sec:conclusion}.
	
	\section{Related works}
	\label{sec:related_work}
	
	Metric learning is an approach based directly on a similarity measure that aims to establish a relationship between data points. The relation provides a semantic similarity such that close data points will be considered similar, whereas remote data points will be considered dissimilar. It can also be categorized as a dimensionality reduction because it maps high-dimensional data points to a low-dimensional space while preserving its separable features \cite{KAMAL2022108562}. Hence, it can also be useful for large-scale data or multimedia applications, which are ubiquitous in the modern era. The popularity of metric learning emerged in 2002 with the pioneering work of Xing et al. \cite{NIPS2002_c3e4035a}, who formulated it as a convex optimization problem. Therefore, several metrics learning applications in various domains have been introduced by some scholars, such as medicine, security, social media mining, speech recognition, information retrieval, recommender systems, and computer vision. In terms of business, the usage of metric learning as a service platform, such as recommendation system and image search similarity, makes consumption and buying decisions more effective and the user experience more comfortable by recommending only relevant items. It allows service providers to predict the customer's usage behavior by reusing professional data mining services.
	
	According to the type of supervision applied during training, the metric learning model mainly falls into two main categories: supervised and weakly supervised learning. In supervised learning, the model is trained using data, where each sample has label information as a standard discriminative model or classifier. In summary, the model learns to map data points based on the label. Thus, the data points in the same label are considered similar and are placed in a close space representation; otherwise, they are mapped in a remote space representation. Popular supervised metric learning models, such as a linear discriminant analysis (LDA), margin maximizing discriminant analysis (MMDA), learning with side information (LSI), relevant component analysis (RCA), and neighborhood component analysis (NCA). Because supervised metric learning directly employs a label in defining the similarity among the data points, the embedding space becomes rigid to the label, is prone to an overfitting similarly to a standard (deep learning) classifier, has difficulty capturing the semantic relationship across data points lying within different categories, and has problems dealing with new (unlisted) categories. In weakly supervised learning, the model has access to a set of data points with supervision at the tuple level (particularly pairs, triplets, quadruplets, or N pairs of data points). Thus, its mechanism is called weakly supervised because the model does not directly map the similarity of data points based on the corresponding label, i.e., the prominent models of weakly supervised metric learning such as Siamese, Triplet, and N-pair networks.
	
	The Siamese network algorithm was first introduced by Bromley et al. to two handwritten signatures in 1994. The model was defined as a binary classifier to distinguish whether the pair of data points belonged to the same class or originated in a different class. Owing to its simplicity and applicability, the Siamese network has become the most widely used network by scholars. Zhang et al. \cite{ZHANG2022105096} developed a content-based image retrieval framework using a Siamese network. In addition, Qiao et al. \cite{8796417} employed a Siamese network to define user identity linkages through web browsing. Moreover, Yang et al. \cite{9052724} presented terahertz image verification using a symmetric Siamese network. As represented by a binary classification problem, the Siamese network solely overlooks the inner-class relationship. Thus, the connection between inter-class data points can be less semantic. To address this issue, the triplet network was introduced by Schroff et al. \cite{7298682} and was originally devised for face recognition. Herein, the data points are mapped in a triplet consisting of an anchor and positive and negative data points; accordingly, the model can learn the similarity of all possible pairs of data points. Some previous triplet network applications are as follows. Zhang et al. \cite{9353191} adopted a triplet loss for remote-sensing image retrieval. Boutros et al. \cite{BOUTROS2022108473} implemented a triplet loss for mask-image recognition. Lue et al. \cite{9376983} also devised a visible-thermal-person re-identification framework using triplet loss. Learning the similarity in a triplet fashion is a promising technique for enhancing the semantic similarity of inter-class structures. Nonetheless, this mechanism is challenging to implement. Many scholars show that a triplet loss often suffers from slow convergence, partially because they employ only one negative example while not interacting with the other negative classes in each update \cite{NIPS2016_6b180037}. Moreover, if some positive and negative data points are similar, the performance can be degraded. Consequently, Sohn and Kihyuk introduced N-pair networks to solve this problem. In an N-pair network, the model learns the pairwise distance or similarity between all possible pairs in N data points. Commonly, N is represented as the number of batches. Therefore, the network is updated by calculating the pairwise similarity within a batch. Among the latest N-pair network applications, Chen and Deng \cite{CHEN2019353} utilized N-pair loss for image retrieval and clustering. In addition, Pal et al. \cite{9388707} employed N-pair loss in biomedical image classification. Moreover, Espejo et al. \cite{9465680} extended the N-pair loss with a $(C_{N,2} + 1)$-pair loss function for keyword spotting. Indeed, an N-pair network is a promising metric learning model for obtaining reliable semantic representations. Nevertheless, the N-pair loss tends to have inferior performance when the number of batches (N) is small. Moreover, employing a large number of batches requires a significantly high computational cost.
	
	In metric learning, the distance or similarity measure also plays a crucial role in defining the embedding space. The base network of a metric learning model commonly produces a vector. Its value must represent the original input both accurately and semantically. The standard distance measure used to define the distance or similarity between vectors is Euclidean and cosine. Nevertheless, these often fail to capture the idiosyncrasies of the data of interest \cite{DBLP:journals/corr/BelletHS13}. Thus, Norouzi et al. \cite{NIPS2012_59b90e10} introduced a framework for learning a broad class of binary hash functions based on a triplet ranking loss designed to preserve the relative similarity. Zhang et al. \cite{NEURIPS2020_d9812f75} proposed a spherical embedding constraint (SEC) to regularize the distribution of the norms. Moreover, some scholars have recently used distance measures to accomplish a better semantic embedding space representation. Xu et al. \cite{NEURIPS2018_814a9c18} used a bi-level distance metric to enhance the similarity accuracy, and Ye et al. \cite{NIPS2016_8fecb208} incorporated multi-metric learning to capture multi-perspective data relationships. Based on the above approaches, we extended this research in another direction. Unlike the previous studies, we devised an expressive but straightforward method for determining the relationship between embedding vectors using covariance metrics.
	
	\begin{figure}[!htbp]
		\centering
		\includegraphics[width=0.9\linewidth]{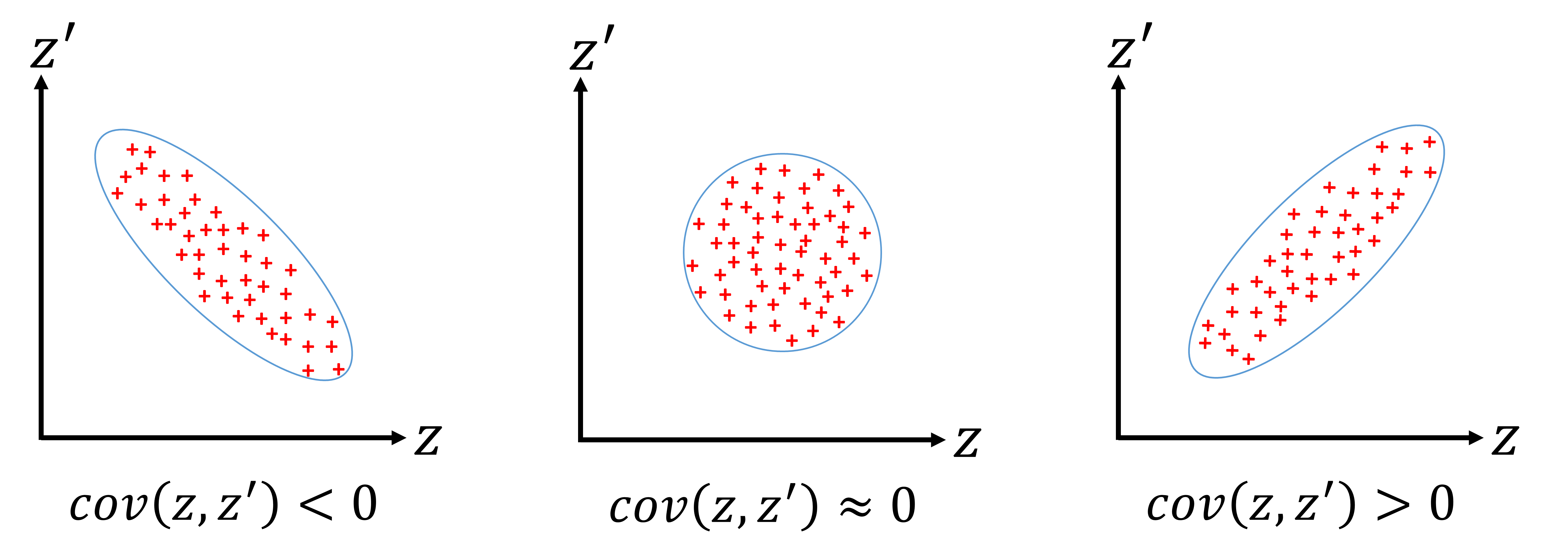}
		\caption{Covariance between two embedding vectors capturing negative, neutral, and positive correlations.}
		\label{fig:covariance_explained}
	\end{figure}
	
	\begin{equation}
		\label{eq:formulation1}
		cov(z,z^{\prime}) = \frac{\sum_{i=1}^{s} (z_i - \bar{z})(z_i^{\prime} - \bar{z}^{\prime})}{s-1}
	\end{equation}
	\begin{equation}
		\label{eq:formulation4}
		\overrightarrow{cov}(z,z^{\prime}) = (z_i - \bar{z})(z_i^{\prime} - \bar{z}^{\prime})
	\end{equation}
	\begin{equation} \label{eq:cov_property}
		\begin{split}
			\sum \overrightarrow{cov}(z,z^{\prime}) 
			\begin{cases}
				> 0, & \textnormal{positive correlation} \\
				\approx 0 , & \textnormal{uncorrelated} \\
				< 0 , & \textnormal{negative correlation} \\
			\end{cases} 
		\end{split}
	\end{equation}
	
	\section{Methodology}
	\label{sec:method}
	
	Rather than Euclidean distance and cosine similarity, the covariance metric is more expressive because it can simultaneously capture three possible relations between two variables, as depicted in Fig. \ref{fig:covariance_explained}. Given two random variables, $z$ and $z^{\prime}$, the relationship between them becomes negative when the $cov(z,z^{\prime}) < 0$, neutral or uncorrelated when $cov(z,z^{\prime}) \approx 0$, and positive when $cov(z,z^{\prime}) > 0$. This value can be extracted from the original covariance equation denoted in Eq. \ref{eq:formulation1}, where $s$ represents the total number of variables, $\bar{z}$ is the mean value of $z$ derived from $\bar{z} = \frac{\sum_{i=1}^{s} z_i}{s}$ and $\bar{z}^{\prime}$ is equal to $\bar{z}^{\prime} = \frac{\sum_{i=1}^{s} z_i^{\prime}}{s}$. Therefore, unlike Euclidean and cosine similarity metrics, the covariance metric is invariant to shifts. If $z$ and $z^{\prime}$ is a vector with $q$ dimension. The covariance between them can be formulated as $cov(z,z^{\prime}) = \sum_{i=1}^{q} (z_i - \bar{z})(z_i^{\prime} - \bar{z}^{\prime})$, where $cov(z,z^{\prime})$ is a scalar value. We represent the covariance vector $\overrightarrow{cov}(z,z^{\prime})$, as shown in Eq. \ref{eq:formulation4}, instead of a scalar because we cannot directly minimize $cov(z,z^{\prime})$ using an optimization algorithm. Note that, the summation of $\overrightarrow{cov}(z,z^{\prime})$ will be positive, neutral (uncorrelated), or negative, as denoted in Eq. \ref{eq:cov_property}.
	
	\subsection{Problem definition}
	
	Let us denote $X = \{x_0,x_1,...,x_{N-1}\}$ as real value $N$ with a $p$-dimensional space, $X \in \mathbb{R}^p$, and $Y = \{y_0, y_1,...,y_{N-1}\}$ as a label of $X$. The embedding network ($F$) will project $X$ into $q$-dimensional space ($z = F(x)$), where $z \in \mathbb{R}^q$ and $q \ll p$. We define a pair input $\{x,x^{\prime}\}$, where $x = x_i$, $x^{\prime} = x_j$, and $i \neq j$. $z^{\prime}$ can also be obtained after $x^{\prime}$ is processed using $F$. Subsequently, we can obtain the covariance value, represented as a vector, between $z$ and $z^{\prime}$ using Eq. \ref{eq:formulation4}. By using a mapping function (IM, IIM, or ISIM), we can obtain a label of $\mathcal{X}=\{x,x^{\prime}\}$, which is represented as $\mathcal{Y}$. Note that $Y$ is a raw label whereas $\mathcal{Y}$ represents the final label generated by the mapping function. The covariance vector $\overrightarrow{cov}(z,z^{\prime})$ is directly projected to become $\mathcal{Y}$ through the function $G$ ($G(\overrightarrow{cov}(z,z^{\prime}))=\mathcal{Y}$), where $G$ represents a dense layer with the number of units $|\mathcal{Y}|$. When the label is binary $|\mathcal{Y}| = 1$, and when the label is categorical $|\mathcal{Y}|$ is equal to the total number of categories. Unlike the Euclidean and cosine similarity metrics, the covariance operation may result in a range value of $[- \infty,\infty]$; hence we cannot directly minimize it by using an optimization algorithm. Thus, we employ a classification approach to implicitly obtain the covariance value between $z$ and $z^{\prime}$.
	
	\begin{figure}[h!]
		\centering
		\subfloat[IM]{\includegraphics[width=0.33333\linewidth]{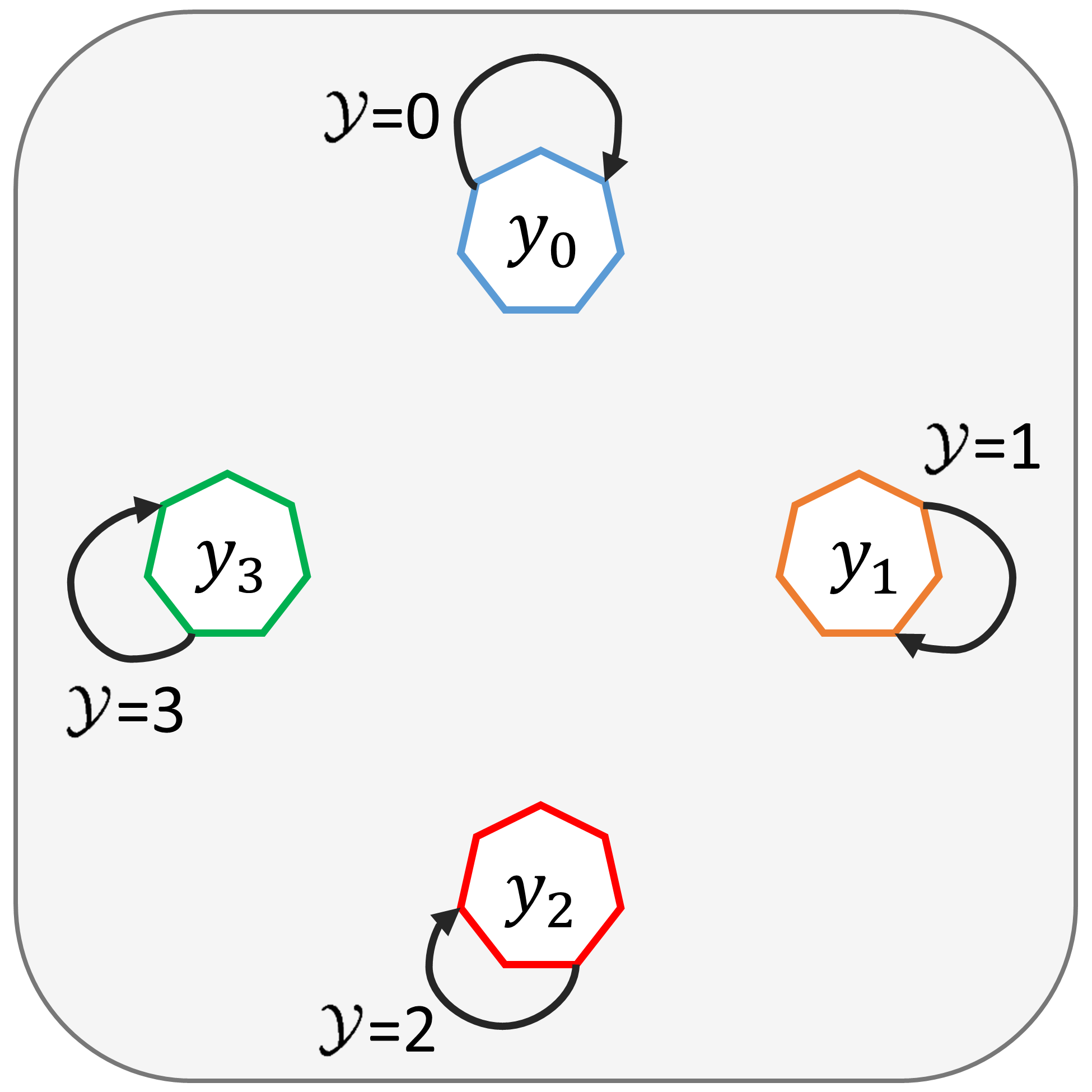}
			\label{strategy_model2}}
		\subfloat[IIM]{\includegraphics[width=0.33333\linewidth]{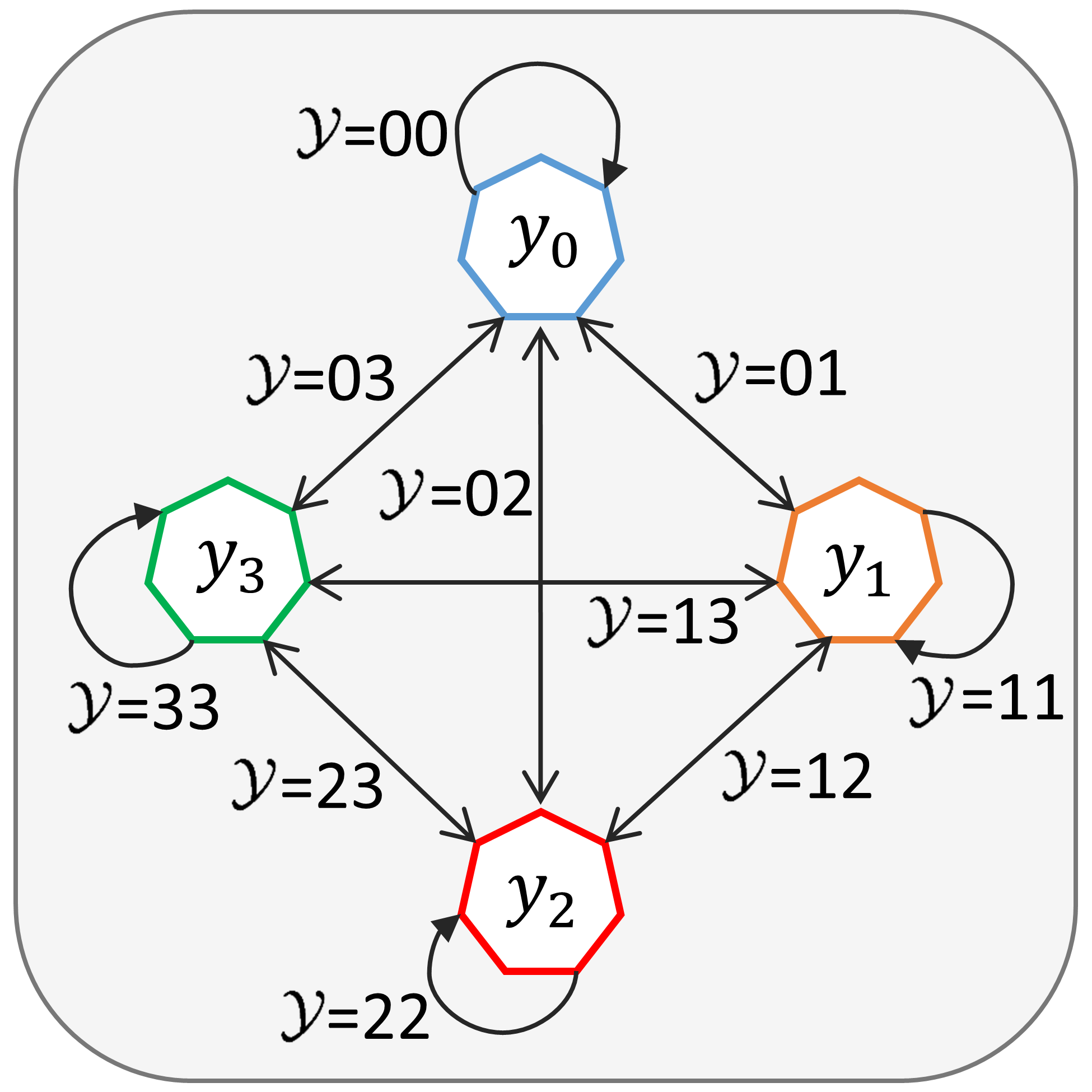}
			\label{strategy_model3}}
		\subfloat[ISIM]{\includegraphics[width=0.33333\linewidth]{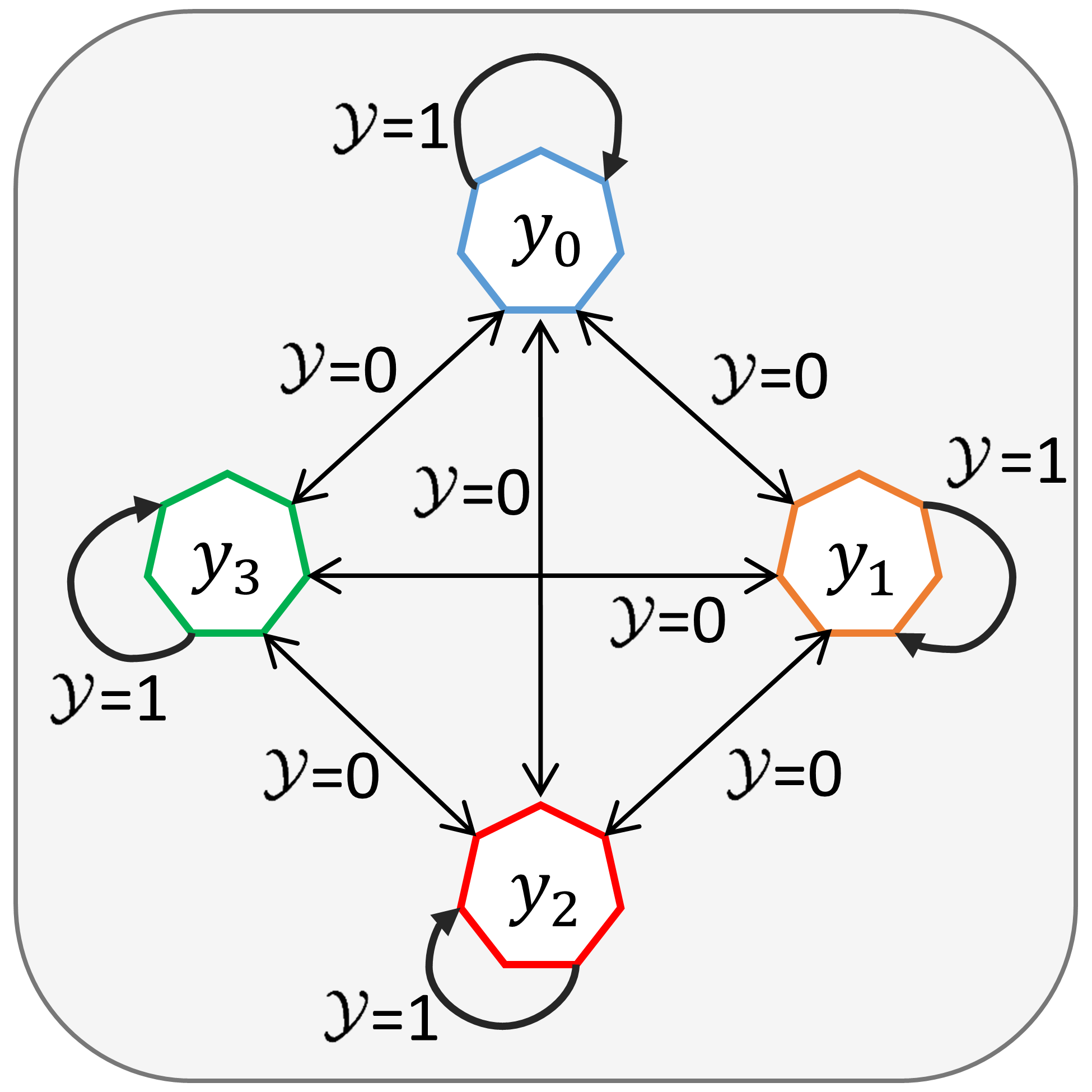}
			\label{strategy_model1}}
		\caption{Mapping function comparison: IM, ISIM, and IIM. Herein there are four categories $\{y_0, y_1,y_2,y_3\}$ and all possible generated labels by mapping function are denoted as $\mathcal{Y}$.}
		\label{fig:data_mapping}
	\end{figure}
	
	\subsection{Data processing and network model}
	
	Before the network model learns the data, we apply a data processing technique which is defined as a mapping function. In this study, we present three types of mapping functions to determine a pair of inputs ($\mathcal{X}=\{x,x^{\prime}\}$) and its label ($\mathcal{Y}$), namely, IM, IIM, and ISIM, as illustrated in Fig. \ref{fig:data_mapping}. Note that $\mathcal{X}$ is an image and $\mathcal{Y}$ is a text. In the IM, the data are paired if they lie in the same class or category. For instance, in the CIFAR-10 dataset, if $x$ is a bulldog, $x^{\prime}$ can be a puddle ($\{x=\text{bulldog},x=\text{puddle}\}, \mathcal{Y}=\text{dog}$) and if $x$ is Persian, $x^{\prime}$ can be a ragdoll ($\{x=\text{Persian},x=\text{ragdoll}\}, \mathcal{Y}=\text{cat}$), and so on. Thus, suppose $C$ is the total number of categories or classes in a dataset, and the total number of classes in IM ($n_{class}$) is equal to the total number of categories in a dataset ($n_{class}$ = 10 in the CIFAR-10 dataset). The detailed IM mapping algorithm is presented in Algorithm \ref{algo:im}. It indicates that the IM explicitly maximizes the inner-class similarity. Nevertheless, it can also implicitly minimize the inter-class similarity after the convergence (when all data points successfully fit with their corresponding categories). In the IIM, the data are paired based on all pair combinations among the categories. For example, ($\{x$=dog, $x^{\prime}$=another dog$\}$, $\mathcal{Y}=\text{dog-dog}$), ($\{x$=dog, $x^{\prime}$=cat$\}$, $\mathcal{Y}=\text{dog-cat}$), ($\{x$=dog, $x^{\prime}$=airplane$\}$, $\mathcal{Y}=\text{dog-airplane}$), ($\{x$=dog, $x^{\prime}$=automobile$\}$, $\mathcal{Y}=\text{dog-automobile}$), ($\{x$=dog, $x^{\prime}$=truck$\}$, $\mathcal{Y}=\text{dog-truck}$), ($\{x$=dog, $x^{\prime}$=ship$\}$, $\mathcal{Y}=\text{dog-ship}$), ($\{x$=dog, $x^{\prime}$=deer$\}$, $\mathcal{Y}=\text{dog-deer}$), ($\{x$=dog, $x^{\prime}$=bird$\}$, $\mathcal{Y}=\text{dog-bird}$), $\{x$=dog, $x^{\prime}$=frog$\}$, $\mathcal{Y}=\text{dog-frog}$), ($\{x$=dog, $x^{\prime}$=horse$\}$, $\mathcal{Y}=\text{dog-horse}$), and so forth for other categories. The total number of classes in IIM can be denoted as $n_{class} = C + \binom{C}{2}$. Because all possible combinations in the class categories are explicitly observed, IIM can learn to maximize and minimize inner- and inter-class similarities simultaneously. The detailed IIM mapping algorithm is presented in Algorithm \ref{algo:iim}. In addition, compared with IM and ISIM, IIM mapping can be robust in extracting the semantic similarity in both inner-class and inter-class relationships because it learns similarity of all combinations of categories. However, because the number of pair combinations can be significantly increased when the number of categories in a dataset is high, it will become a classification with a large number of categories. In which, this is still an active area of research in machine learning \cite{ABRAMOVICH2019104536}. In ISIM, the data are paired in the same as in the original Siamese network mapping. Therefore, there are only two possible labels (binary) for ISIM, namely, a pair containing the same class category (e.g. $\{x$=dog, $x^{\prime}$=another dog$\}, \mathcal{Y} = 1$), and a pair consisting of a different class category (e.g. $\{x$=dog, $x^{\prime}$=cat$\}, \mathcal{Y} = 0$). Because it is assumed that the similarity between a dog and a cat is equal to that between a dog and an airplane, the capturing of the inter-class relationship may be explicitly neglected. The detailed ISIM mapping is presented in Algorithm \ref{algo:isim}.
	
	\begin{figure}[h!]
		\centering
		\includegraphics[width=0.45\linewidth]{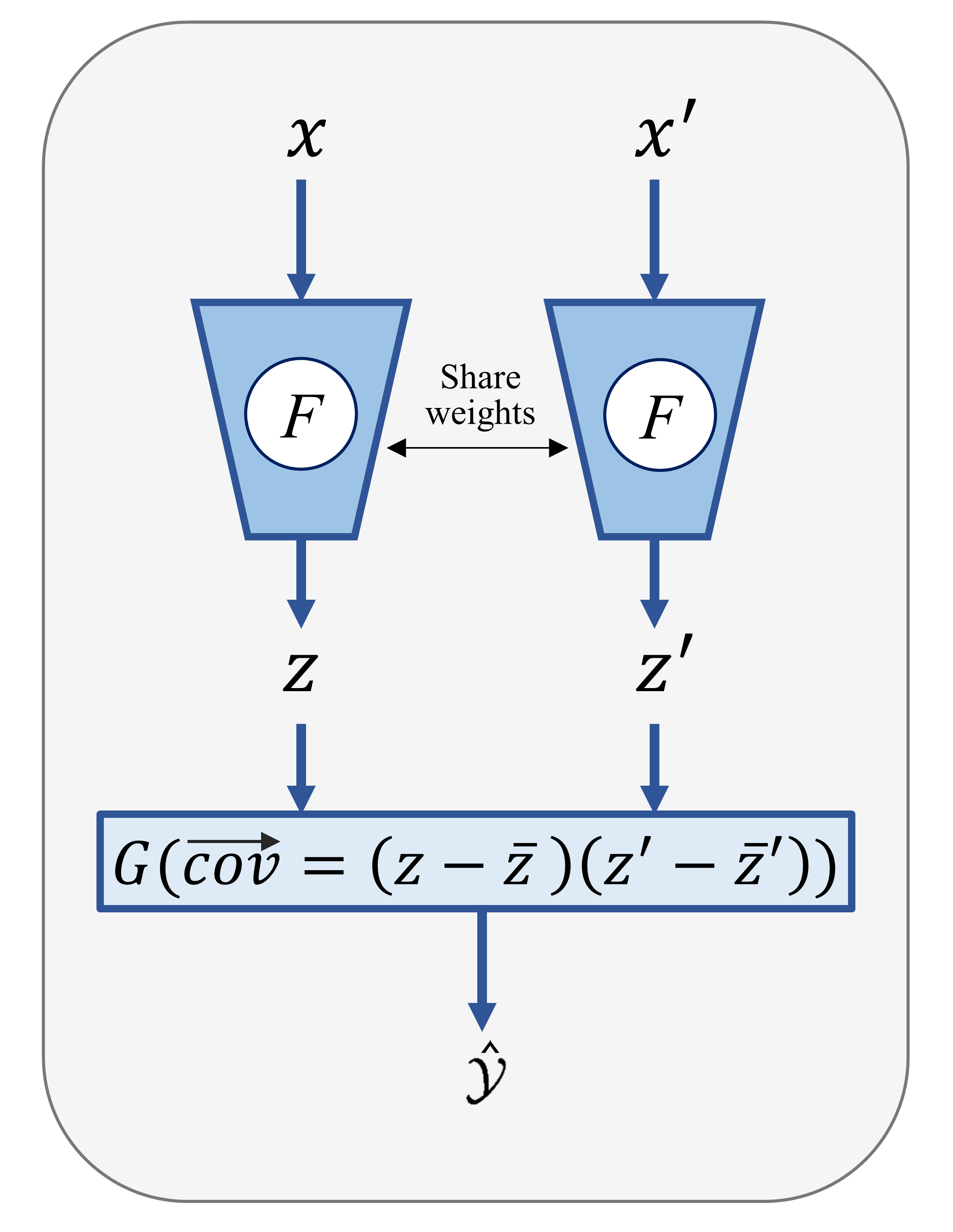}
		\caption{Covariance network (CovNet). Herein the input $\{x,x^{\prime}\}$ and output $\mathcal{Y}$ are obtained from the mapping function.}
		\label{fig:proposed_model}
	\end{figure}
	
	\begin{algorithm}[h!]
		\caption{IM}
		\setstretch{0.75}
		\SetAlgoLined
		\label{algo:im}
		\KwIn{$X=$\{$x_0,x_1,...,x_{N-1}$\}, $Y=$\{$y_0,y_1,...,x_{N-1}$\}}
		\KwOut{$\mathcal{X}$, $\mathcal{Y}$} 
		$CG$ \tcp*{In $CG$, $x$ is grouped and indexed by $y$}
		$\mathcal{X}$ = [] \tcp*{initialize $\mathcal{X}$ as an empty list}
		$\mathcal{Y}$ = [] \tcp*{initialize $\mathcal{Y}$ as an empty list}
		\For{$i \gets 0$ \KwTo $N-1$}{
			\Do{$x_i = x_j$}{
				$x_j \gets$ \text{\textbf{random}}($CG_{y_i}$) \tcp*{get random $x_j$ in $CG$, where $y_j=y_i$}
			}
			$\mathcal{X}.$\text{\textbf{append}}($\{x_i,x_j\}$) \tcp*{pair $x$ and $x^{\prime}$}
			$\mathcal{Y}.$\text{\textbf{append}}($y_i$) \tcp*{label, note that $y_i = y_j$}
		}
	\end{algorithm}
	
	\begin{algorithm}[h!]
		\setstretch{0.75}
		\SetAlgoLined
		\caption{IIM}
		\label{algo:iim}
		\KwIn{$X=$\{$x_0,x_1,...,x_{N-1}$\}, $Y=$\{$y_0,y_1,...,x_{N-1}$\}}
		\KwOut{$\mathcal{X}$, $\mathcal{Y}$} 
		$CG$ \tcp*{In $CG$, $x$ is grouped and indexed by $y$}
		$\mathcal{X}$ = [] \tcp*{initialize $\mathcal{X}$ as an empty list}
		$\mathcal{Y}$ = [] \tcp*{initialize $\mathcal{Y}$ as an empty list}
		\For{$i \gets 0$ \KwTo $N-1$}{
			\For{$j \gets 0$ \KwTo $C-1$}{
				\uIf{$y_i = y_j$}{ 
					\Do{$x_i = x_j$}{
						$x_j \gets$ \text{\textbf{random}}($CG_{y_i}$) \tcp*{get random $x_j$ in $CG$,where $y_j=y_i$}
					}
				}
				\Else{
					$x_j \gets$ \text{\textbf{random}}($CG_{y_j}$) \tcp*{get random $x_j$ in $CG$, where $y_j \neq y_i$}
				}
				$\mathcal{X}.$\text{\textbf{append}}($\{x_i,x_j\}$) \tcp*{pair $x$ and $x^{\prime}$}
				\uIf{$i \leq j$}{
					$\mathcal{Y}.$\text{\textbf{append}}(\textbf{str}($y_i$)+\textbf{str}($y_j$)) \tcp*{label as a string}
				}
				\Else{
					$\mathcal{Y}.$\text{\textbf{append}}(\textbf{str}($y_j$)+\textbf{str}($y_i$)) \tcp*{label as a string in ascending order only}
				}
			}
			
		}
	\end{algorithm}
	
	\begin{algorithm}[h!]
		\setstretch{0.75}
		\SetAlgoLined
		\caption{ISIM}
		\label{algo:isim}
		\KwIn{$X=$\{$x_0,x_1,...,x_{N-1}$\}, $Y=$\{$y_0,y_1,...,x_{N-1}$\}}
		\KwOut{$\mathcal{X}$, $\mathcal{Y}$} 
		$CG$ \tcp*{In $CG$, $x$ is grouped and indexed by $y$}
		$\mathcal{X}$ = [] \tcp*{initialize $\mathcal{X}$ as an empty list}
		$\mathcal{Y}$ = [] \tcp*{initialize $\mathcal{Y}$ as an empty list}
		\For{$i \gets 0$ \KwTo $N-1$}{
			\tcc{add a matching pair}
			\Do{$x_i = x_j$}{
				$x_j \gets$ \text{\textbf{random}}($CG_{y_i}$) \tcp*{get random $x_j$ in $CG$, where $y_j=y_i$}
			}
			$\mathcal{X}.$\text{\textbf{append}}($\{x_i,x_j\}$) \tcp*{ pair $x$ and $x^{\prime}$}
			$\mathcal{Y}.$\text{\textbf{append}}(1) \tcp*{a matching label, $\mathcal{Y}=1$}
			\tcc{add a non-matching pair}
			\Do{$y_j = y_i$}{
				$y_j \gets$ \text{\textbf{random\_int}}($0,C-1$) \tcp*{to make sure $y_i \neq y_j$}
			}
			$x_j \gets$ \text{\textbf{random}}($CG_{y_j}$) \tcp*{get random $x_j$ in $CG$ where $y_j \neq y_i$}
			$\mathcal{X}.$\text{\textbf{append}}($\{x_i,x_j\}$) \tcp*{pair $x$ and $x^{\prime}$}
			$\mathcal{Y}.$\text{\textbf{append}}(0) \tcp*{a non-matching label, $\mathcal{Y}=0$}
		}
	\end{algorithm}
	
	The data format obtained from the processing technique is a pair input, $\{x,x^{\prime}\}$ with a single label, $\mathcal{Y}$. They were trained in a supervised manner using a network model called CovNet. Thus, $\{x,x^{\prime}\}$ is processed to estimate the label $\mathcal{\hat{Y}}$. The overall architecture of CovNet is illustrated in Fig. \ref{fig:proposed_model}. We define $F$ as an embedding network represented as a convolutional neural network (CNN). We visually present two embedding networks. However, because they share their weights, they are physically a single network ($F$). As a standard CNN, the model consists of convolutional, max-pooling, batch normalization, and dropout layers. In the last layer, we employ a global average-pooling layer to capture the extracted feature generated from the convolutional layer. Subsequently, a dense layer with L2 normalization (L2 norm.) is applied to obtain the embedding vector ($z$). The L2 norm. has also been commonly used in previous metric learning models because it can result in stability during training. Thus, because $z$ and $z^{\prime}$ are already normalized through the L2 norm, if the $z$ and $z^{\prime}$ are centered (have zero means), it (Eq. \ref{eq:formulation4}) will have the same result as the cosine similarity. In the tail layer, we incorporate it with a dense layer ($G$) using a softmax ($S$) or sigmoid function to determine the class probability. 
	
	\subsection{Learning mechanism}
	
	CovNet was trained as a standard neural network with a back-propagation algorithm. The learning mechanism of CovNet is described in Algorithm \ref{algo:covnet}. The inputs are $\mathcal{X}$ and $\mathcal{Y}$, the values of which are obtained from the mapping function used (IM, IIM, or ISIM, as defined in Algorithms \ref{algo:im}, \ref{algo:iim}, and \ref{algo:isim}, respectively). Note that each of the samples in $\mathcal{X}$ consists of tuples $\{x,x^{\prime}\}$, and $\mathcal{Y}$ is represented as a one-hot vector (in IM and IIM mapping) and a binary vector (in ISIM mapping). The output of the algorithm is the optimum embedding network model ($F^*_{{\phi}}$). Initially, $F_{\phi}$ produces $z$ and $z^{\prime}$ given input $x$ and $x^{\prime}$, respectively. Subsequently, we can obtain $\overrightarrow{cov}(z,z^{\prime})$, which is formulated in Eq. \ref{eq:formulation4}. Then, $G_{\psi}$ will classify whether $\overrightarrow{cov}(z,z^{\prime})$ agrees on the same label. A predefined \textit{loss\_function} calculates the discrepancy loss between $\mathcal{Y}$ and $\mathcal{\hat{Y}}$. If we employ IM and IIM as our mapping function, the \textit{loss\_function} is the categorical cross-entropy ($\mathcal{L}_{CE}$) formulated in Eqs. \ref{eq:formulation_softmax} and \ref{eq:formulation_ce}. However, if we utilize ISIM as our mapping function, the \textit{loss\_function} is a binary cross-entropy ($\mathcal{L}_{BE}$), denoted in Eq. \ref{eq:formulation_be}. For simplicity, the model parameters $F$($\phi$) and $G$ ($\psi$) are wrapped as a single parameter $\Omega$, and will be updated using a gradient-descent based optimization algorithm. All aforementioned processes were repeated until reaching the predefined number of epochs ($NoEpoch$). Note that, in practice, the model is trained and updated using a predefined number of batches.
	
	\begin{equation}
		\label{eq:formulation_softmax}
		S(\mathcal{\hat{Y}}_i) = \frac{{\exp}^{\mathcal{\hat{Y}}_i}}{\sum_{i=1}^{n_{class}}\exp^{\mathcal{\hat{Y}}_i}}
	\end{equation}
	\begin{equation}
		\label{eq:formulation_ce}
		\mathcal{L}_{CE} = - \sum_{i=1}^{n_{class}} {\mathcal{Y}}_i \log(S(\mathcal{\hat{Y}}_i))
	\end{equation}
	\begin{equation}
		\label{eq:formulation_be}
		\mathcal{L}_{BE} = - [\mathcal{Y} \log(\mathcal{\hat{Y}}) + (1-\mathcal{Y}) \log(1-\mathcal{\hat{Y}})]
	\end{equation}
	
	\begin{algorithm}[h!]
		\setstretch{0.75}
		\SetAlgoLined
		\caption{training procedure of CovNet}
		\label{algo:covnet}
		\KwIn{$\mathcal{X}$ = \{$\{x_0,x^{\prime}_0\},\{x_1,x^{\prime}_1\},...,\{x_{N-1},x^{\prime}_{N-1}\}$\}, 
			$\mathcal{Y}$ = \{$\mathcal{Y}_0,\mathcal{Y}_1,...,\mathcal{Y}_{N-1}$\}} 
		\KwOut{$F^*_{{\phi}}$}
		\For{$i \gets 1$ \KwTo $NoEpoch$}{
			$z = F_{{\phi}_i}(x)$ \tcp*{embedding network extracts $z$}
			$z^{\prime} = F_{{\phi}_i}(x^{\prime})$ \tcp*{embedding network extracts $z^{\prime}$}
			
			$\overrightarrow{cov}(z,z^{\prime}) =$ Eq. \ref{eq:formulation4} \tcp*{extract covariance vector}
			$\mathcal{\hat{Y}} = G_{{\psi}_i}(\overrightarrow{cov}(z,z^{\prime}))$\tcp*{estimate $\mathcal{Y}$}
			$\mathcal{L}_{{\Omega}_i}({\phi}_i, {\psi}_i) =$\text{\textbf{loss\_function}($\mathcal{Y},\mathcal{\hat{Y}}$)} \tcp*{calculate supervised loss betwen $\mathcal{Y}$ and $\mathcal{\hat{Y}}$}
			${\Omega}_i \gets {\Omega}_i - \eta \nabla_{{\Omega}_i}{\mathcal{L}}_{{\Omega}_i}$ \tcp*{update model parameter, $\Omega = \{\phi, \psi\}$}
		}
	\end{algorithm}
	
	\begin{figure*}[h!]
		\centering
		\includegraphics[width=0.95\linewidth]{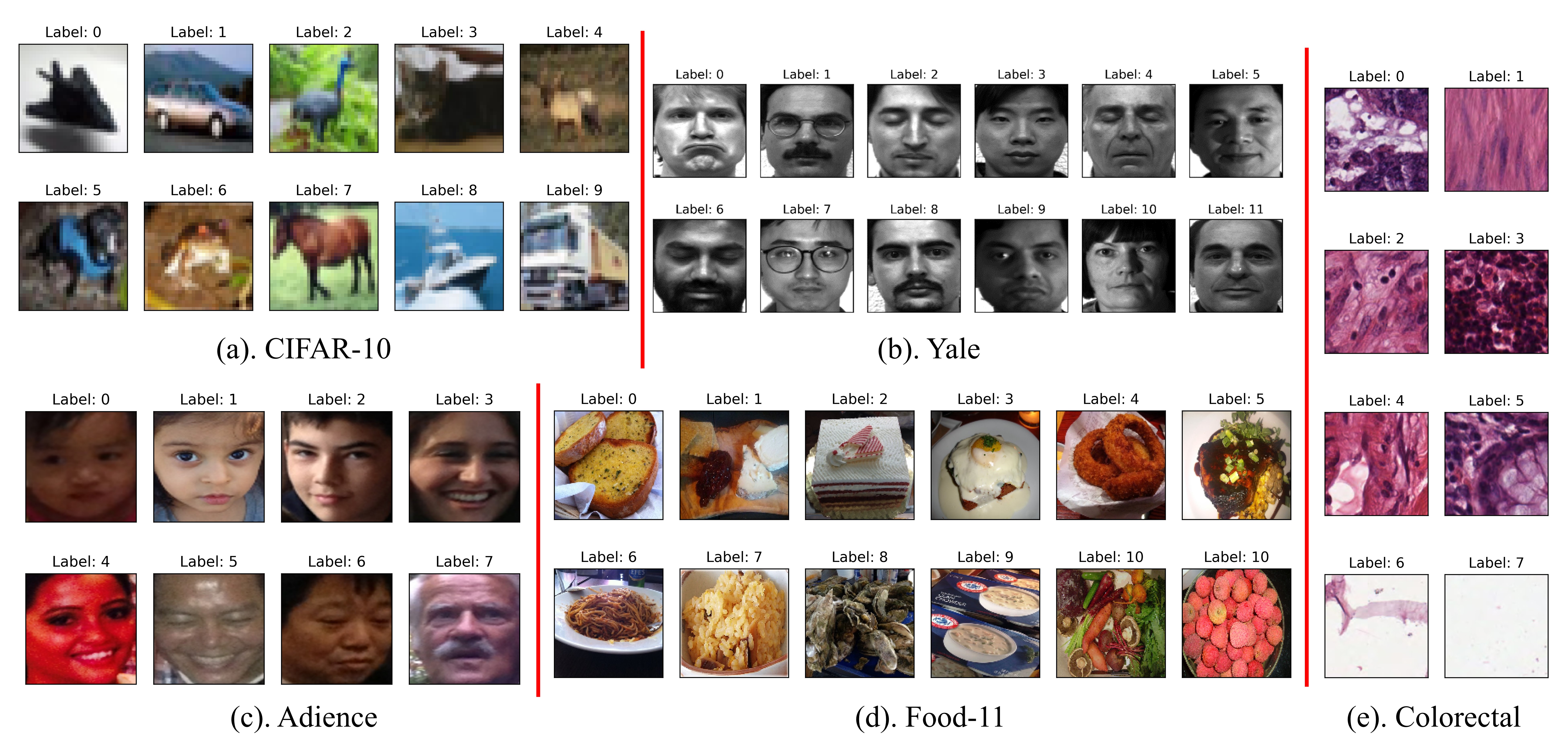}
		\caption{Dataset overview}
		\label{fig:dataset_overview}
	\end{figure*}
	
	\section{Experiments}
	\label{sec:experiment}
	
	This section presents the performance evaluation of our proposed model (CovNets), followed by a comparison and discussion between CovNets and existing metric learning models.
	
	\begin{table*}[!t]
		\centering
		\caption{Embedding network ($F$) of CovNet, Siamese, Triplet, and N-pair networks in all datasets.}
		\label{tab:architecture}
		\resizebox{0.95\linewidth}{!}{
			\begin{tabular}{l}		
				\toprule
				\textbf{CIFAR-10}:			\\
				$Conv(64,3,Relu)-BN-Conv(64,3,Relu)-MP(3)-BN-DO(0.5)-Conv(128,3,Relu)-BN-$ \\
				$Conv(128,3,Relu)-MP(3)-BN-DO(0.5)-Conv(256,3,Relu)-BN-Conv(256,3,Relu)-$ \\
				$Conv(256,3,Relu)-MP(3)-BN-DO(0.5)-AP-Dense(256,Relu)-BN-DO(0.5)-Dense(100,Tanh)-L2Norm.$ \\
				\midrule
				\textbf{Food-11 \& Colorectal}: 	\\
				$Conv(64,3,Relu)-BN-Conv(64,3,Relu)-MP(3)-BN-DO(0.5)-Conv(128,3,Relu)-BN-$\\
				$Conv(128,3,Relu)-MP(3)-BN-DO(0.5)-Conv(256,3,Relu)-BN-Conv(256,3,Relu)-$\\
				$Conv(512,3,Relu)-MP(3)-BN-DO(0.5)-AP-Dense(256,Relu)-BN-DO(0.5)-Dense(100,Tanh)-L2Norm.$ \\
				\midrule
				\textbf{Adience}: 	\\
				$ZP(5)-FaceNet-Dense(512,Relu)-BN-DO(0.25)-Dense(100,Tanh)-L2Norm.$ \\
				\midrule
				\textbf{Yale}: 	\\
				$FaceNet-Dense(512,Relu)-BN-DO(0.25)-Dense(100,Tanh)-L2Norm.$ \\
				\bottomrule
			\end{tabular}
		}
	\end{table*}
	
	\subsection{Experimental setting}
	\label{sec:experiment_setting}
	
	We evaluated our model on various well-known datasets including those with natural images (CIFAR-10 \cite{Krizhevsky09learningmultiple} and Food-11 \cite{10.1145/2986035.2986039}), biomedical images (Colorectal \cite{kather2016multi}), and facial images (Adience \cite{6906255} and Yale \cite{1247387}). Sample images from each dataset are illustrated in Fig. \ref{fig:dataset_overview}. The CIFAR-10 dataset consists of 50,000 $32 \times 32$ pixels color training images and 10,000 test images, labeled over ten categories: airplanes, automobiles, birds, cats, deer, dogs, frogs, horses, ships, and trucks. We obtained 10\% of the training set for the validation set. Food-11 is a dataset containing 16,643 food images grouped into 11 major food categories: bread, dairy products, desserts, eggs, fried foods, meat, noodles/pasta, rice, seafood, soup, and vegetable/fruit. The original images had various dimensions and were split into 9,866 training, 3,430 validation, and 3,347 testing sets. For simplicity, we uniformly resized the image dimensions to 128$\times$128 pixels. We employ zero-contrast normalization and ZCA whitening as data preprocessing on Food-11 and CIFAR-10. Colorectal cancer is a biomedical dataset containing histological tiles from patients with colorectal cancer. It is made up of 150×150 pixels color images from eight classes, i.e., debris, mucosa, tumor, adipose, stroma, lympho, complex, and empty. The original dataset consisted of 5,000 samples. We randomly split the datasets into training, validation, and testing sets at a ratio of 70\%, 10\%, and 20\%, respectively. We reshaped the images into 128$\times$128 pixels, standardized, and normalized to a value range of 0–1. Adience is composed of face images scraped from \textit{Flickr.com} albums that were labeled for age and gender. The benchmark uses eight classes for age groups (0–2, 4–6, 8–13, 15–20, 25–32, 38–43, 48–53, 60+). A total of 38,740 images were split into five groups of 4,484, 3,730, 3,894, 3,446, 3,816, and 19,370. The Yale dataset contains 165 grayscale images of 15 individuals in GIF format. There were 11 images per subject, 1 for each different facial expression or configuration: center-light, w/glasses, happy, left-light, without glasses, normal, right-light, sad, sleepy, surprised, and wink. We preprocessed the Yale and Adience datasets by cropping and centering on the faces and resized them into 150$\times$150 pixels and 160$\times$160 pixels color images for the Adience and Yale datasets, respectively. In addition, we conducted five cross-validations to assess the model performance.
	
	\begin{table*}[!t]
		\centering
		\caption{Comparison of network architectures.}
		\label{tab:total_model_comp}
		\resizebox{0.78\linewidth}{!}{
			\begin{threeparttable}
				\centering
				\begin{tabular}{l l l l l l}		
					\toprule
					Model							&	Input mapping		&	Embeding net.					&	Merging layer						&	Tail layer							&	Loss function	 			\\
					\midrule
					\multirow{2}{*}{CovNet v1}		&	\multirow{2}{*}{IM}	&	\multirow{2}{*}{$F_{dataset}$}	&	\multirow{2}{*}{Covariance} 		&	Dense($n_{class}$,					&	Categorical					\\[-1pt]
					&						&									&										&	Softmax)							&	crossentropy				\\
					\midrule
					\multirow{2}{*}{CovNet v2}		&	\multirow{2}{*}{IIM}&	\multirow{2}{*}{$F_{dataset}$}	&	\multirow{2}{*}{Covariance} 		&	Dense($n_{class}$,					&	Categorical					\\[-1pt]
					&						&									&			 							&	Softmax)							&	crossentropy				\\
					\midrule
					\multirow{2}{*}{CovNet v3}		&	\multirow{2}{*}{ISIM}&	\multirow{2}{*}{$F_{dataset}$}	&	\multirow{2}{*}{Covariance}  		&	BN + Dense(1,						&	Binary 						\\[-1pt]
					&						&									&			 							&	Sigmoid)							&	crossentropy				\\
					\midrule
					\multirow{2}{*}{Siamese}		&	\multirow{2}{*}{ISIM}&	\multirow{2}{*}{$F_{dataset}$}	&	\multirow{2}{*}{Euclidean dist.} 	&	BN + Dense(1,						&	Contrastive						\\[-1pt]
					&						&									&					 					&	Sigmoid)							&	loss				\\
					\midrule
					Triplet							&	TM\tnote{*}					&	$F_{dataset}$					&	Triplet dist. 						&	--									&	Triplet loss				\\
					\midrule
					\multirow{2}{*}{N-pair}			&	\multirow{2}{*}{IM}	&	\multirow{2}{*}{$F_{dataset}$}	&	\multirow{2}{*}{--}					&	\multirow{2}{*}{--}					&	N-pair cosine				\\[-1pt]
					&						&									&										&										&	similarity					\\
					\bottomrule
				\end{tabular}
				\begin{tablenotes}
					\item[*] Triplet mapping.
				\end{tablenotes}
			\end{threeparttable}
		}
	\end{table*}

	We compared our experimental results with the state-of-art metric learning models, such as Siamese, Triplet, and N-pair networks, to assess the effectiveness of our proposed model. We employed the same embedding network ($F$) in all models for a fair comparison, as shown in Table \ref{tab:architecture}. Here, $Conv(i,j,k)$ denotes a convolutional layer with $i$ number of filters, a kernel size of $j \times j$, and $k$ activation functions. In addition, $MP(i)$ is a 2D max-pooling layer with a size of $i \times i$, $BN$ is a batch normalization layer, and $DO(i)$ corresponds to the dropout layer with probability $i$. In addition, $AP$ is a 2D average-pooling layer, $Dense(i, j)$ represents a dense layer with $i$ neurons and $j$ activation functions. Moreover, $ZP(i)$ indicates zero-padding with size $i$. For the Adience and Yale datasets, we employed a pre-trained network (FaceNet) \cite{7298682}, which is an inception model trained on the MS-Celeb-1M dataset, as a backbone network.
	
	The overall network architecture of all models is summarized in Table \ref{tab:total_model_comp}. The value of $F$ can vary based on the dataset, which is referred to in Table \ref{tab:architecture}. Here, CovNet v1 and CovNet v2 have different properties in both the input mapping and tail layer. In this case, CovNet v1 utilizes IM, whereas CovNet v2 employs IIM, which is notably more complex than IM. Because $n_{class}$ is defined by all pair combinations of the class label in CovNet v2, the $n_{class}$ of the CovNet v2 value is significantly higher than that of CovNet v1. The $n_{class}$ of CovNet v1 is equal to the total number of class categories. In addition, CovNet v3 has the same architecture as the Siamese network except in the merging layer, which is represented as a binary classification. Both triplet and N-pair networks are more straightforward than the others because they do not have a tail layer. Nevertheless, their learning mechanism can be trickier because triplet and N-pair losses are directly optimized during the training phase. We employed batch sizes of 128, 128, 32, 64, and 16 for the CIFAR-10, Food-11, Colorectal, Adience, and Yale datasets, respectively. Each network was trained for 200 epochs using Adam optimization. An early stopping mechanism was executed when the best model was obtained based on the validation sets. 
	\begin{equation}
		\label{eq:accuracy_knn}
		accuracy = \frac{\sum_{i=1}^{N} \sum_{j=1}^{k} c_i^j}{N \times k}
	\end{equation}
	\begin{equation} \label{eq:accuracy_knn2}
		\begin{split}
			c_i^j 
			\begin{cases}
				= 1, & \mathcal{\hat{Y}}_i^j = \mathcal{Y}_i \\
				0 , & \text{otherwise} \\
			\end{cases} 
		\end{split}
	\end{equation}
	\begin{figure*}[!t]
		\centering
		\begin{tikzpicture}[scale=0.27]
			\begin{axis}[
				title style={at={(1.2,4)},anchor=south,yshift=-0.1},
				title={CIFAR-10},
				ylabel={Accuracy},
				set layers,
				ybar=1.2pt,
				width  = 0.88*\textwidth,
				bar width=8pt,
				symbolic x coords={CovNet-v1,CovNet-v2,CovNet-v3,Siamese,Triplet,N-pair,N-pair(1000)},
				samples=7,
				ymax=0.82,
				xtick=data,
				x label style={font=\small},
				y label style={font=\small},
				ticklabel style={font=\small},
				xticklabel style={rotate=45, anchor = east},
				scaled y ticks = false,
				y tick label style={/pgf/number format/fixed},
				yticklabel style={/pgf/number format/fixed,/pgf/number format/precision=2},
				legend image code/.code={%
					\draw[#1, draw=none] (0cm,-0.1cm) rectangle (0.275cm,0.07cm);
				},  
				legend style={at={(1.2,3.6)},
					anchor=south,legend columns=-1},
				every axis plot/.append style={fill opacity=0.6},
				every tick label/.append style={font=\small},
				log origin y=infty
				]
				\addplot[black,fill=red!50,postaction={pattern=north east lines},error bars/.cd,
				y dir=both,y explicit] coordinates {
					(CovNet-v1, 0.79881) +- (0.0, 0.005)
					(CovNet-v2, 0.81125) +- (0.0, 0.003)
					(CovNet-v3, 0.79947) +- (0.0, 0.006)
					(Siamese, 0.79702) +- (0.0, 0.003)
					(Triplet, 0.68925) +- (0.0, 0.006)
					(N-pair, 0.62844) +- (0.0, 0.007)
					(N-pair(1000), 0.77325) +- (0.0, 0.005)
				};
				\addlegendentry{$k$=10}
				\addplot[black,fill=yellow!50,postaction={pattern=north west lines},error bars/.cd,
				y dir=both,y explicit] coordinates {
					(CovNet-v1, 0.79594875) +- (0.0, 0.006)
					(CovNet-v2, 0.8100325) +- (0.0, 0.004)
					(CovNet-v3, 0.79768625) +- (0.0, 0.006)
					(Siamese, 0.79493) +- (0.0, 0.004)
					(Triplet, 0.6794125) +- (0.0, 0.009)
					(N-pair, 0.62612) +- (0.0, 0.009)
					(N-pair(1000), 0.76667) +- (0.0, 0.007)
				};
				\addlegendentry{$k$=40}
				\addplot[black,fill=green!50,postaction={pattern=dots},error bars/.cd,
				y dir=both,y explicit] coordinates {
					(CovNet-v1, 0.79398642) +- (0.0, 0.006)
					(CovNet-v2, 0.800857) +- (0.0, 0.005)
					(CovNet-v3, 0.7952857) +- (0.0, 0.006)
					(Siamese, 0.7930161) +- (0.0, 0.005)
					(Triplet, 0.67362499) +- (0.0, 0.008)
					(N-pair, 0.62521071) +- (0.0, 0.008)
					(N-pair(1000), 0.760637142857142) +- (0.0, 0.006)
				};
				\addlegendentry{$k$=70}
				\addplot[black,fill=blue!50,postaction={pattern=crosshatch},error bars/.cd,
				y dir=both,y explicit] coordinates {
					(CovNet-v1, 0.7936885) +- (0.0, 0.006)
					(CovNet-v2, 0.799672) +- (0.0, 0.005)
					(CovNet-v3, 0.7903012) +- (0.0, 0.006)
					(Siamese, 0.7901961) +- (0.0, 0.005)
					(Triplet, 0.666098) +- (0.0, 0.006)
					(N-pair, 0.6201865) +- (0.0, 0.009)
					(N-pair(1000), 0.752276) +- (0.0, 0.008)
				};
				\addlegendentry{$k$=100}
			\end{axis}
		\end{tikzpicture} \hspace{1.5mm}
		\begin{tikzpicture}[scale=0.27]
			\begin{axis}[
				title style={at={(1.2,4)},anchor=south,yshift=-0.1},
				title={Food-11},
				ylabel={Accuracy},
				set layers,
				ybar=1.2pt,
				width  = 0.88*\textwidth,
				bar width=8pt,
				symbolic x coords={CovNet-v1,CovNet-v2,CovNet-v3,Siamese,Triplet,N-pair, N-pair(1000)},
				samples=7,
				ymax=0.78,
				x label style={font=\small},
				y label style={font=\small},
				ticklabel style={font=\small},
				xticklabel style={rotate=45, anchor = east},
				xtick=data,
				scaled y ticks = false,
				y tick label style={/pgf/number format/fixed},
				yticklabel style={/pgf/number format/fixed,/pgf/number format/precision=2},
				legend image code/.code={%
					\draw[#1, draw=none] (0cm,-0.1cm) rectangle (0.3cm,0.07cm);
				},  
				legend style={at={(1.2,3.6)},
					anchor=south,legend columns=-1},
				every axis plot/.append style={fill opacity=0.6},
				every tick label/.append style={font=\small},
				log origin y=infty
				]
				\addplot[black,fill=red!50,postaction={pattern=north east lines},error bars/.cd,
				y dir=both,y explicit] coordinates {
					(CovNet-v1, 0.74393486) +- (0.0, 0.007)
					(CovNet-v2, 0.7591398267) +- (0.0, 0.006)
					(CovNet-v3, 0.73956827009262) +- (0.0, 0.006)
					(Siamese, 0.744158948311921) +- (0.0, 0.005)
					(Triplet, 0.18164027487302) +- (0.0, 0.01)
					(N-pair, 0.38054974604123) +- (0.0, 0.009)
					(N-pair(1000), 0.654167911562593) +- (0.0, 0.006)
				};
				\addlegendentry{$k$=10}
				\addplot[black,fill=yellow!50,postaction={pattern=north west lines},error bars/.cd,
				y dir=both,y explicit] coordinates {
					(CovNet-v1, 0.741219749) +- (0.0, 0.007)
					(CovNet-v2, 0.7587182551) +- (0.0, 0.007)
					(CovNet-v3, 0.736786674634) +- (0.0, 0.005)
					(Siamese, 0.741992829399462) +- (0.0, 0.006)
					(Triplet, 0.178163280549746) +- (0.0, 0.008)
					(N-pair, 0.373819838661487) +- (0.0, 0.009)
					(N-pair(1000), 0.649208246190618) +- (0.0, 0.007)
				};
				\addlegendentry{$k$=40}
				\addplot[black,fill=green!50,postaction={pattern=dots},error bars/.cd,
				y dir=both,y explicit] coordinates {
					(CovNet-v1, 0.7365764650) +- (0.0, 0.008)
					(CovNet-v2,0.758175765077) +- (0.0, 0.006)
					(CovNet-v3, 0.736111229672628) +- (0.0, 0.006)
					(Siamese, 0.738200520722181) +- (0.0, 0.006)
					(Triplet, 0.175730931751248) +- (0.0, 0.009)
					(N-pair, 0.370860898885996) +- (0.0, 0.01)
					(N-pair(1000), 0.646391224550778) +- (0.0, 0.005)
				};
				\addlegendentry{$k$=70}
				\addplot[black,fill=blue!50,postaction={pattern=crosshatch},error bars/.cd,
				y dir=both,y explicit] coordinates {
					(CovNet-v1, 0.7297669554) +- (0.0, 0.008)
					(CovNet-v2, 0.75700627427) +- (0.0, 0.006)
					(CovNet-v3, 0.732118314908873) +- (0.0, 0.007)
					(Siamese, 0.730902300567672) +- (0.0, 0.007)
					(Triplet, 0.173721242904092) +- (0.0, 0.009)
					(N-pair, 0.366982372273677) +- (0.0, 0.01)
					(N-pair(1000), 0.642103376157753) +- (0.0, 0.008)
				};
				\addlegendentry{$k$=100}
			\end{axis}
		\end{tikzpicture}\hspace{1.5mm}
		\begin{tikzpicture}[scale=0.27]
			\begin{axis}[
				title style={at={(1.2,4)},anchor=south,yshift=-0.1},
				title={Colorectal},
				ylabel={Accuracy},
				set layers,
				ybar=1.2pt,
				width  = 0.88*\textwidth,
				bar width=8pt,
				symbolic x coords={CovNet-v1,CovNet-v2,CovNet-v3,Siamese,Triplet,N-pair, N-pair(1000)},
				samples=7,
				ymax=1,
				x label style={font=\small},
				y label style={font=\small},
				xtick=data,
				ticklabel style={font=\small},
				xticklabel style={rotate=45, anchor = east},
				scaled y ticks = false,
				y tick label style={/pgf/number format/fixed},
				yticklabel style={/pgf/number format/fixed,/pgf/number format/precision=2},
				legend image code/.code={%
					\draw[#1, draw=none] (0cm,-0.1cm) rectangle (0.3cm,0.07cm);
				},  
				legend style={at={(1.2,3.6)},
					anchor=south,legend columns=-1},
				every axis plot/.append style={fill opacity=0.6},
				every tick label/.append style={font=\small},
				log origin y=infty
				]
				\addplot[black,fill=red!50,postaction={pattern=north east lines},error bars/.cd,
				y dir=both,y explicit] coordinates {
					(CovNet-v1, 0.95875) +- (0.0, 0.005)
					(CovNet-v2, 0.9908) +- (0.0, 0.003)
					(CovNet-v3, 0.9677) +- (0.0, 0.005)
					(Siamese, 0.94585) +- (0.0, 0.006)
					(Triplet, 0.6128) +- (0.0, 0.006)
					(N-pair, 0.9228) +- (0.0, 0.004)
					(N-pair(1000), 0.9795)+- (0.0, 0.006)
				};
				\addlegendentry{$k$=10}
				\addplot[black,fill=yellow!50,postaction={pattern=north west lines},error bars/.cd,
				y dir=both,y explicit] coordinates {
					(CovNet-v1, 0.9544375) +- (0.0, 0.006)
					(CovNet-v2, 0.9906) +- (0.0, 0.005)
					(CovNet-v3, 0.965375) +- (0.0, 0.006)
					(Siamese, 0.9399875) +- (0.0, 0.006)
					(Triplet, 0.611425) +- (0.0, 0.007)
					(N-pair, 0.920325) +- (0.0, 0.007)
					(N-pair(1000), 0.97865) +- (0.0, 0.006)
				};
				\addlegendentry{$k$=40}
				\addplot[black,fill=green!50,postaction={pattern=dots},error bars/.cd,
				y dir=both,y explicit] coordinates {
					(CovNet-v1, 0.948) +- (0.0, 0.005)
					(CovNet-v2, 0.989521428571428) +- (0.0, 0.005)
					(CovNet-v3, 0.957414285714286) +- (0.0, 0.006)
					(Siamese, 0.93855) +- (0.0, 0.006)
					(Triplet, 0.584185714285714) +- (0.0, 0.007)
					(N-pair, 0.915542857142857) +- (0.0, 0.008)
					(N-pair(1000),0.972185714285714) +- (0.0, 0.005)
				};
				\addlegendentry{$k$=70}
				\addplot[black,fill=blue!50,postaction={pattern=crosshatch},error bars/.cd,
				y dir=both,y explicit] coordinates {
					(CovNet-v1, 0.942325) +- (0.0, 0.006)
					(CovNet-v2, 0.98937) +- (0.0, 0.004)
					(CovNet-v3, 0.95684) +- (0.0, 0.005)
					(Siamese, 0.93808) +- (0.0, 0.005)
					(Triplet, 0.57941) +- (0.0, 0.005)
					(N-pair, 0.913985) +- (0.0, 0.007)
					(N-pair(1000),0.96705) +- (0.0, 0.006)
				};
				\addlegendentry{$k$=100}
			\end{axis}
		\end{tikzpicture}		
		
		\begin{tikzpicture}[scale=0.27]
			\begin{axis}[
				title style={at={(1.35,4)},anchor=south,yshift=-0.1},
				title={Adience},
				ylabel={Accuracy},
				set layers,
				ybar=1.2pt,
				width  = 0.88*\textwidth,
				bar width=8pt,
				symbolic x coords={CovNet-v1,CovNet-v2,CovNet-v3,Siamese,Triplet,N-pair,N-pair(1000)},
				samples=7,
				ymax=0.84,
				x label style={font=\small},
				y label style={font=\small},
				xtick=data,
				ticklabel style={font=\small},
				xticklabel style={rotate=45, anchor = east},
				scaled y ticks = false,
				y tick label style={/pgf/number format/fixed},
				yticklabel style={/pgf/number format/fixed,/pgf/number format/precision=2},
				legend image code/.code={%
					\draw[#1, draw=none] (0cm,-0.1cm) rectangle (0.3cm,0.07cm);
				},  
				legend style={at={(1.35,3.6)},
					anchor=south,legend columns=-1},
				every axis plot/.append style={fill opacity=0.6},
				every tick label/.append style={font=\small},
				log origin y=infty
				]
				\addplot[black,fill=red!50,postaction={pattern=north east lines},error bars/.cd,
				y dir=both,y explicit] coordinates {
					(CovNet-v1, 0.798511668460322)+- (0.0, 0.007)
					(CovNet-v2, 0.83177425606238) +- (0.0, 0.005)
					(CovNet-v3, 0.817344322500504) +- (0.0, 0.007)
					(Siamese, 0.803499109162633) +- (0.0, 0.005)
					(Triplet, 0.796261517792391) +- (0.0, 0.008)
					(N-pair, 0.790183502212752) +- (0.0, 0.008)
					(N-pair(1000),0.805517220781773) +- (0.0, 0.004)
				};
				\addlegendentry{$k$=2}
				\addplot[black,fill=yellow!50,postaction={pattern=north west lines},error bars/.cd,
				y dir=both,y explicit] coordinates {
					(CovNet-v1, 0.73924793383035) +- (0.0, 0.007)
					(CovNet-v2, 0.770663992079858) +- (0.0, 0.006)
					(CovNet-v3, 0.753173730658989) +- (0.0, 0.007)
					(Siamese, 0.731714506768554) +- (0.0, 0.007)
					(Triplet, 0.732794151985499) +- (0.0, 0.007)
					(N-pair, 0.732632437558783) +- (0.0, 0.005)
					(N-pair(1000),0.747122789374149) +- (0.0, 0.005)
				};
				\addlegendentry{$k$=6}
				\addplot[black,fill=green!50,postaction={pattern=dots},error bars/.cd,
				y dir=both,y explicit] coordinates {
					(CovNet-v1, 0.706238532472241) +- (0.0, 0.004)
					(CovNet-v2, 0.735148762996655) +- (0.0, 0.004)
					(CovNet-v3, 0.716306609077417) +- (0.0, 0.006)
					(Siamese, 0.688321242141874) +- (0.0, 0.007)
					(Triplet, 0.70131011997293) +- (0.0, 0.008)
					(N-pair, 0.703148334264632) +- (0.0, 0.005)
					(N-pair(1000),0.715008693483845) +- (0.0, 0.006)
				};
				\addlegendentry{$k$=10}
				\addplot[black,fill=blue!50,postaction={pattern=crosshatch},error bars/.cd,
				y dir=both,y explicit] coordinates {
					(CovNet-v1, 0.684017277934692) +- (0.0, 0.006)
					(CovNet-v2, 0.70993695909223) +- (0.0, 0.005)
					(CovNet-v3, 0.690213890673491) +- (0.0, 0.006)
					(Siamese, 0.657678272352652) +- (0.0, 0.006)
					(Triplet, 0.681180393388925) +- (0.0, 0.005)
					(N-pair, 0.683012917229322) +- (0.0, 0.005)
					(N-pair(1000),0.693106878730164) +- (0.0, 0.007)
				};
				\addlegendentry{$k$=14}
			\end{axis}
		\end{tikzpicture} \hspace{1.5mm}
		\begin{tikzpicture}[scale=0.27]
			\begin{axis}[
				title style={at={(1.4,4)},anchor=south,yshift=-0.1},
				title={Yale},
				ylabel={Accuracy},
				set layers,
				ybar=1.3pt,
				width  = 0.88*\textwidth,
				bar width=8pt,
				symbolic x coords={CovNet-v1,CovNet-v2,CovNet-v3,Siamese,Triplet,N-pair,N-pair(1000)},
				samples=7,
				ymax=1.01,
				x label style={font=\small},
				y label style={font=\small},
				xtick=data,
				ticklabel style={font=\small},
				xticklabel style={rotate=45, anchor = east},
				scaled y ticks = false,
				y tick label style={/pgf/number format/fixed},
				yticklabel style={/pgf/number format/fixed,/pgf/number format/precision=2},
				legend image code/.code={%
					\draw[#1, draw=none] (0cm,-0.1cm) rectangle (0.3cm,0.07cm);
				},  
				legend style={at={(1.45,3.6)},
					anchor=south,legend columns=-1},
				every axis plot/.append style={fill opacity=0.6},
				every tick label/.append style={font=\small},
				log origin y=infty
				]
				\addplot[black,fill=red!50,postaction={pattern=north east lines},error bars/.cd,
				y dir=both,y explicit] coordinates {
					(CovNet-v1, 1) +- (0.0, 0.005)
					(CovNet-v2, 1) +- (0.0, 0.003)
					(CovNet-v3, 1) +- (0.0, 0.004)
					(Siamese, 0.99) +- (0.0, 0.004)
					(Triplet, 1) +- (0.0, 0.006)
					(N-pair, 1) +- (0.0, 0.005)
					(N-pair(1000),1) +- (0.0, 0.004)
				};
				\addlegendentry{$k$=1}
				\addplot[black,fill=yellow!50,postaction={pattern=north west lines},error bars/.cd,
				y dir=both,y explicit] coordinates {
					(CovNet-v1, 1) +- (0.0, 0.006)
					(CovNet-v2, 1) +- (0.0, 0.004)
					(CovNet-v3, 0.994166633266667) +- (0.0, 0.006)
					(Siamese, 0.974166666666663) +- (0.0, 0.005)
					(Triplet, 0.995833333333333) +- (0.0, 0.005)
					(N-pair, 1) +- (0.0, 0.006)
					(N-pair(1000),1) +- (0.0, 0.004)
				};
				\addlegendentry{$k$=2}
				\addplot[black,fill=green!50,postaction={pattern=dots},error bars/.cd,
				y dir=both,y explicit] coordinates {
					(CovNet-v1, 1) +- (0.0, 0.005)
					(CovNet-v2, 1) +- (0.0, 0.004)
					(CovNet-v3, 0.9811110777) +- (0.0, 0.006)
					(Siamese, 0.942777777777777) +- (0.0, 0.006)
					(Triplet, 0.983333333333333) +- (0.0, 0.007)
					(N-pair, 1) +- (0.0, 0.005)
					(N-pair(1000),1) +- (0.0, 0.006)
				};
				\addlegendentry{$k$=3}
				\addplot[black,fill=blue!50,postaction={pattern=crosshatch},error bars/.cd,
				y dir=both,y explicit] coordinates {
					(CovNet-v1, 0.75) +- (0.0, 0.005)
					(CovNet-v2, 0.75) +- (0.0, 0.005)
					(CovNet-v3, 0.747916666633333) +- (0.0, 0.006)
					(Siamese, 0.716666666666666) +- (0.0, 0.006)
					(Triplet,0.743333332999997) +- (0.0, 0.007)
					(N-pair, 0.75) +- (0.0, 0.005)
					(N-pair(1000),0.75) +- (0.0, 0.004)
				};
				\addlegendentry{$k$=4}
			\end{axis}
			
		\end{tikzpicture}
		\caption{Effect of varying $k$ (nearest neighbor) on image search accuracyy.}\label{fig:comparison_knn}
	\end{figure*}
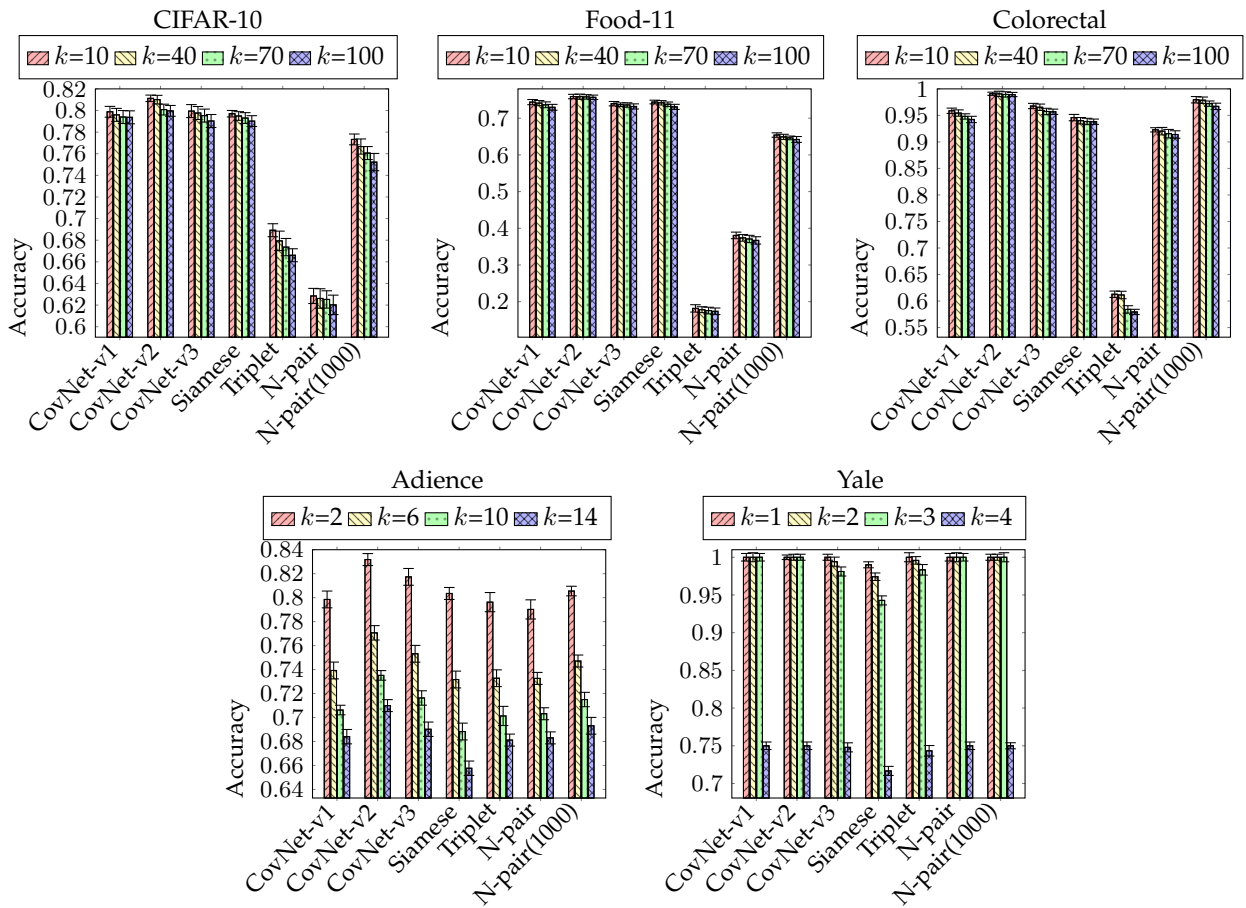
	\subsection{Near neighbor analysis}
	The outcome of an embedding network is commonly represented as a vector having a significantly lower dimension than the original input. The main objective of the embedding method is to preserve the data separability in the embedding space to be the same as possible as in the original space. Accordingly, any machine learning task with complex and large dimensional datasets, such as images, can be simply solved using the near neighbor algorithm. Equation \ref{eq:accuracy_knn} measures the accuracy of the near neighbors of each point in the embedding space based on the corresponding label, where $k$ represents the number of neighbors, and $N$ is the total number of data. The $c_i^j$ value is 1 when the data point and the predicted neighbor have the same label; otherwise, it is zero, as shown in Eq. \ref{eq:accuracy_knn2}. Note that the denominator of Eq. \ref{eq:accuracy_knn} is $N \times k$, which represents the total number of neighbors of each sample in the dataset. In this study, we employed a simple $k$-nearest neighbor algorithm with cosine similarity to measure the closeness among the points (represented as vectors) in the embedding space.
	
	The $k$ nearest neighbor performance of the embedding network for all models is shown in Fig. \ref{fig:comparison_knn}. In general, the accuracy decreases slightly with increasing values of $k$ (except for Adience, in which the accuracy is significantly decreased). This indicates that the higher the value of $k$ is, the more diverse the neighbors are. Nevertheless, in real applications, such as search engines and recommendation systems, users are commonly interested only in the top-10 neighbors. CovNet v2 outperforms other methods in all datasets because it explicitly minimizes the inner-class separability and maximizes the inter-class diversity using IIM. A Siamese network is the most competitive existing method for the proposed models. However, it has a slightly lower performance compared to the other approaches for the Adience and Yale datasets, which used a pre-trained network. In comparison, both Siamese and CovNet v3 utilized binary cross-entropy, which directly neglected the inter-class relationship. However, CovNet v3 is generally better than a Siamese network. Thus, we can reveal that employing covariance in the merging layer is more robust than the Euclidean distance. The triplet network for both Food-11 and Colorectal has worse performance than the other models. This is because some samples in different classes can be nearly similar, as shown in Fig. \ref{fig:dataset_overview}; for example, images labeled 3 and 6 in Food-11 are on the same plate, and those labeled 6 and 7 in colorectal have nearly similar characteristics. Accordingly, the distance between anchor-positive and anchor-negative can be inconstant. In addition, the N-pair with a small batch achieves a low performance. By contrast, if we increase the number of batches to 1000 (N-pair(1000)), its performance significantly increases. However, the computational cost is considerably increased. Finally, we can observe that if we utilize pre-trained networks (Adience and Yale) as a backbone, the performances of all models are nearly competitive with each other.
	
\begin{figure*}[!t]
		\centering
		\subfloat[CovNet v1]{\includegraphics[width=0.28\linewidth]{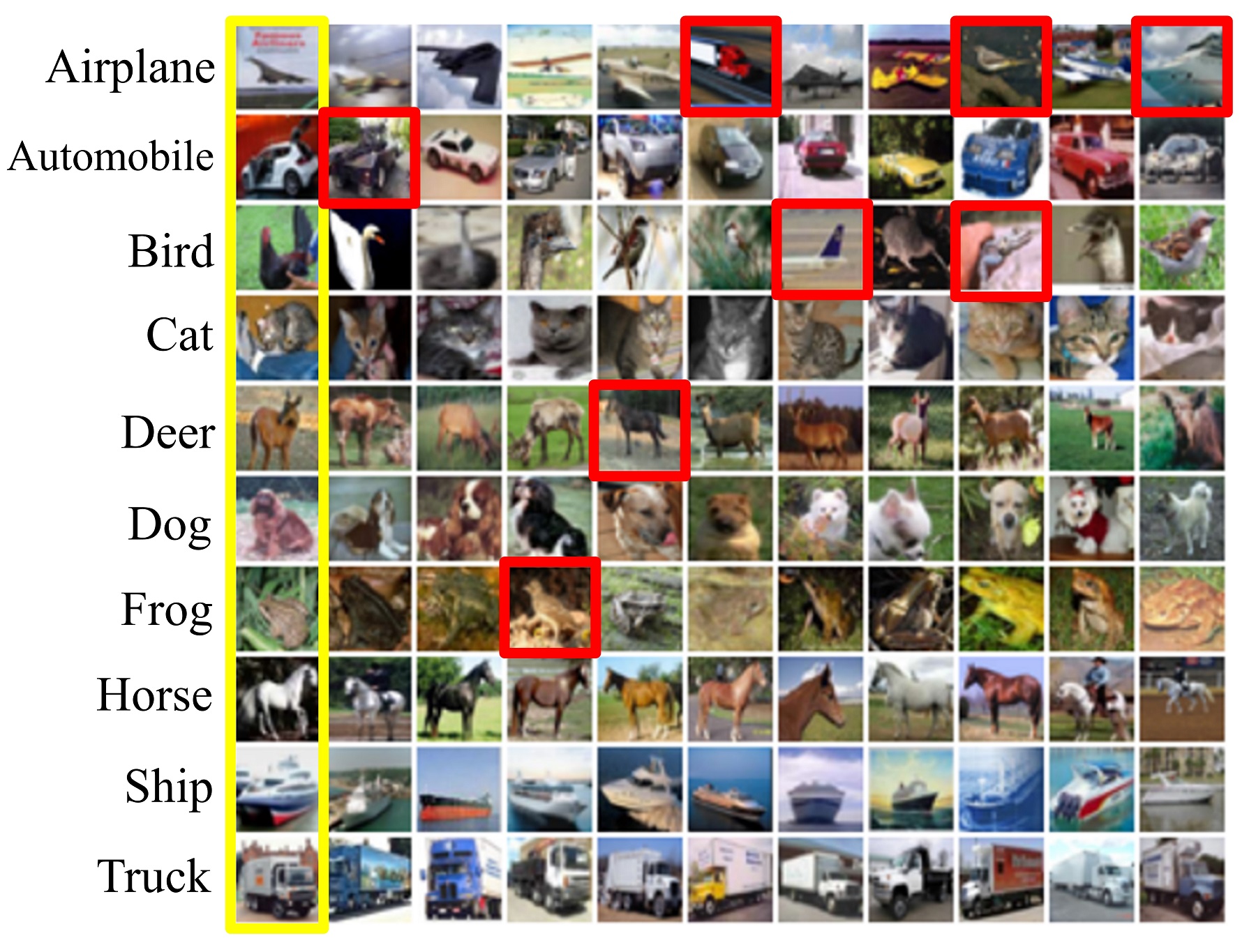}
			\label{search_cifar_v1}}
		\subfloat[CovNet v2]{\includegraphics[width=0.28\linewidth]{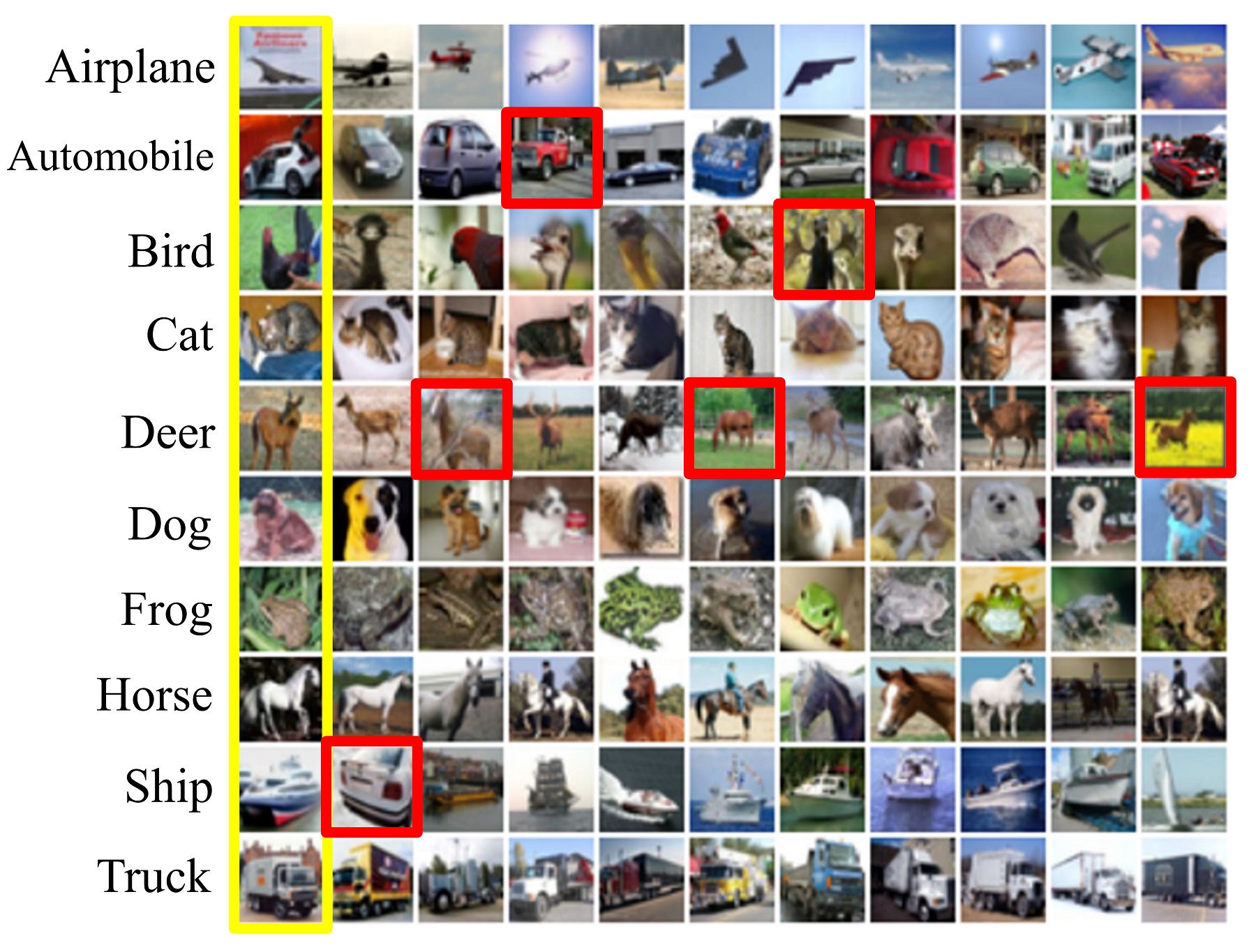}
			\label{search_cifar_v2}}
		\subfloat[CovNet v3]{\includegraphics[width=0.28\linewidth]{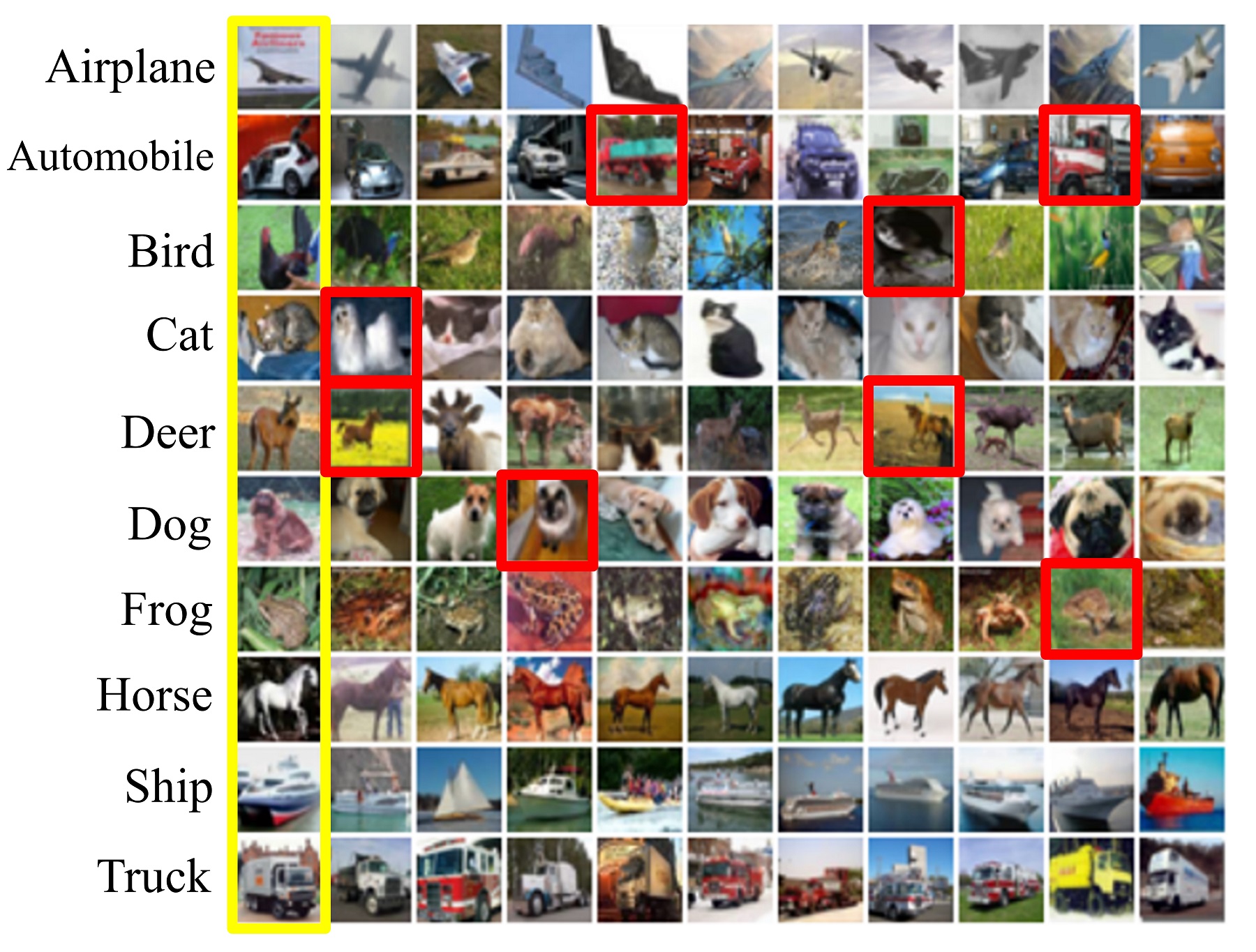}
			\label{search_cifar_v3}}
		
		\subfloat[Siamese]{\includegraphics[width=0.28\linewidth]{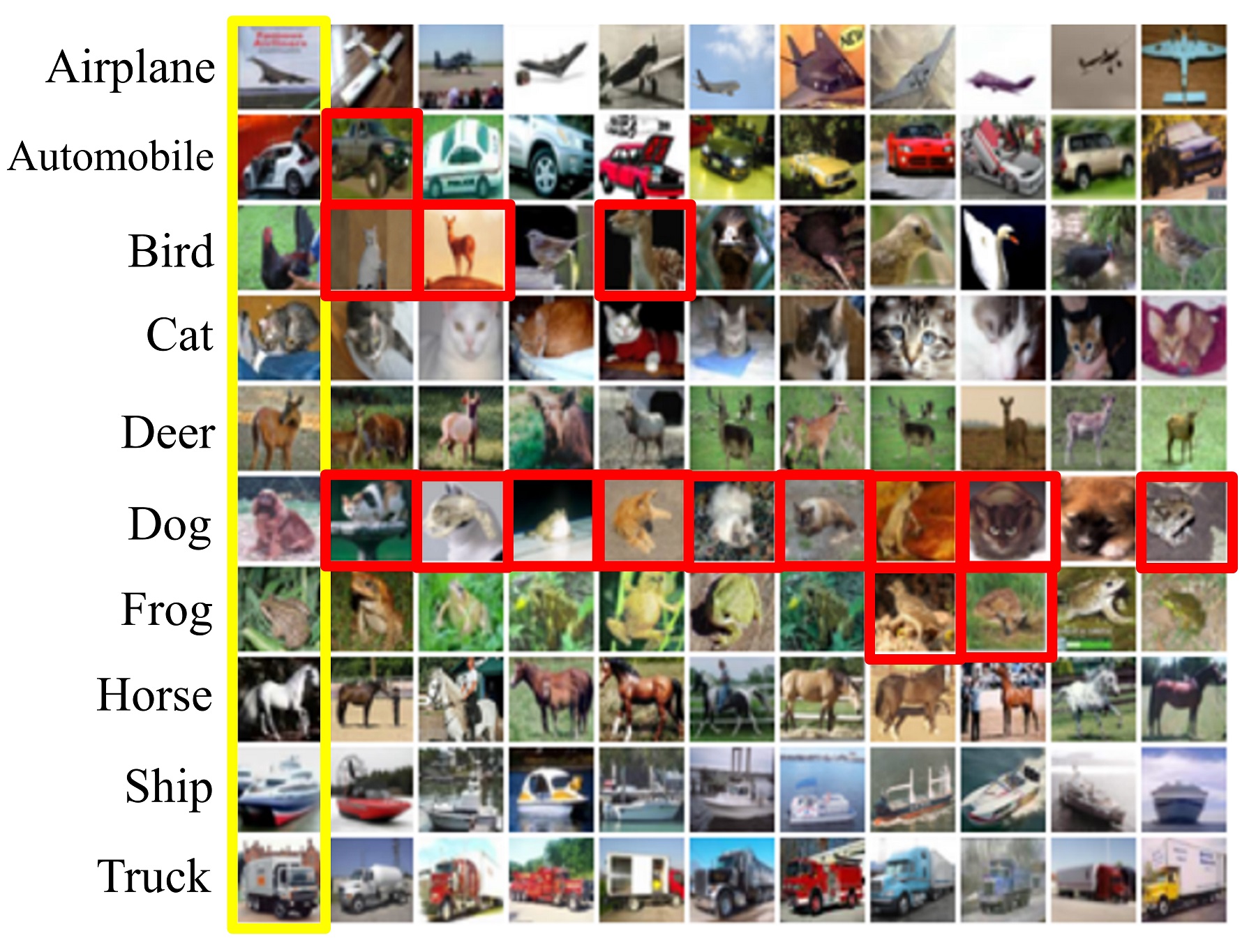}
			\label{search_cifar_siamese}}
		\subfloat[Triplet]{\includegraphics[width=0.28\linewidth]{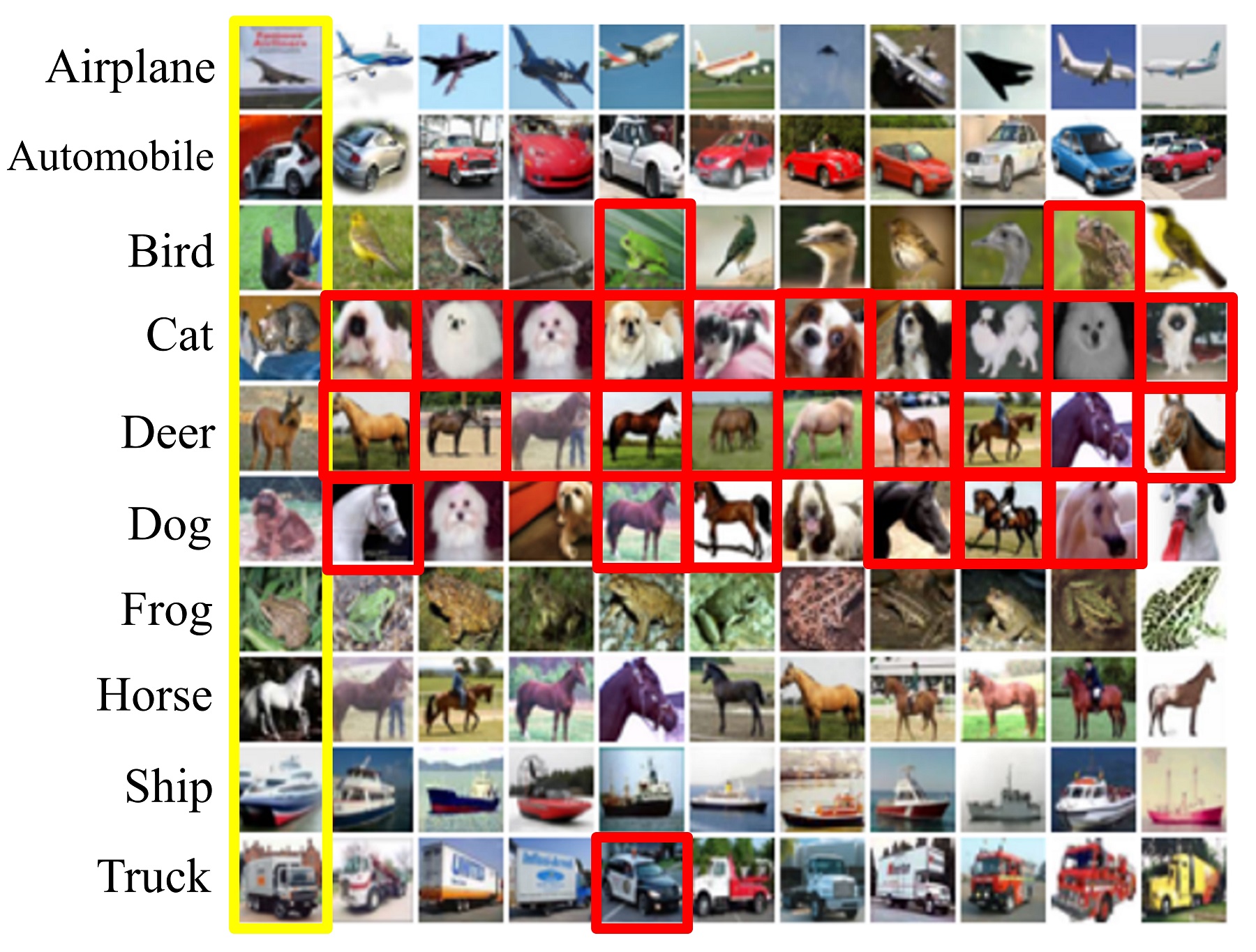}
			\label{search_cifar_triplet}}
		\subfloat[N-Pair]{\includegraphics[width=0.28\linewidth]{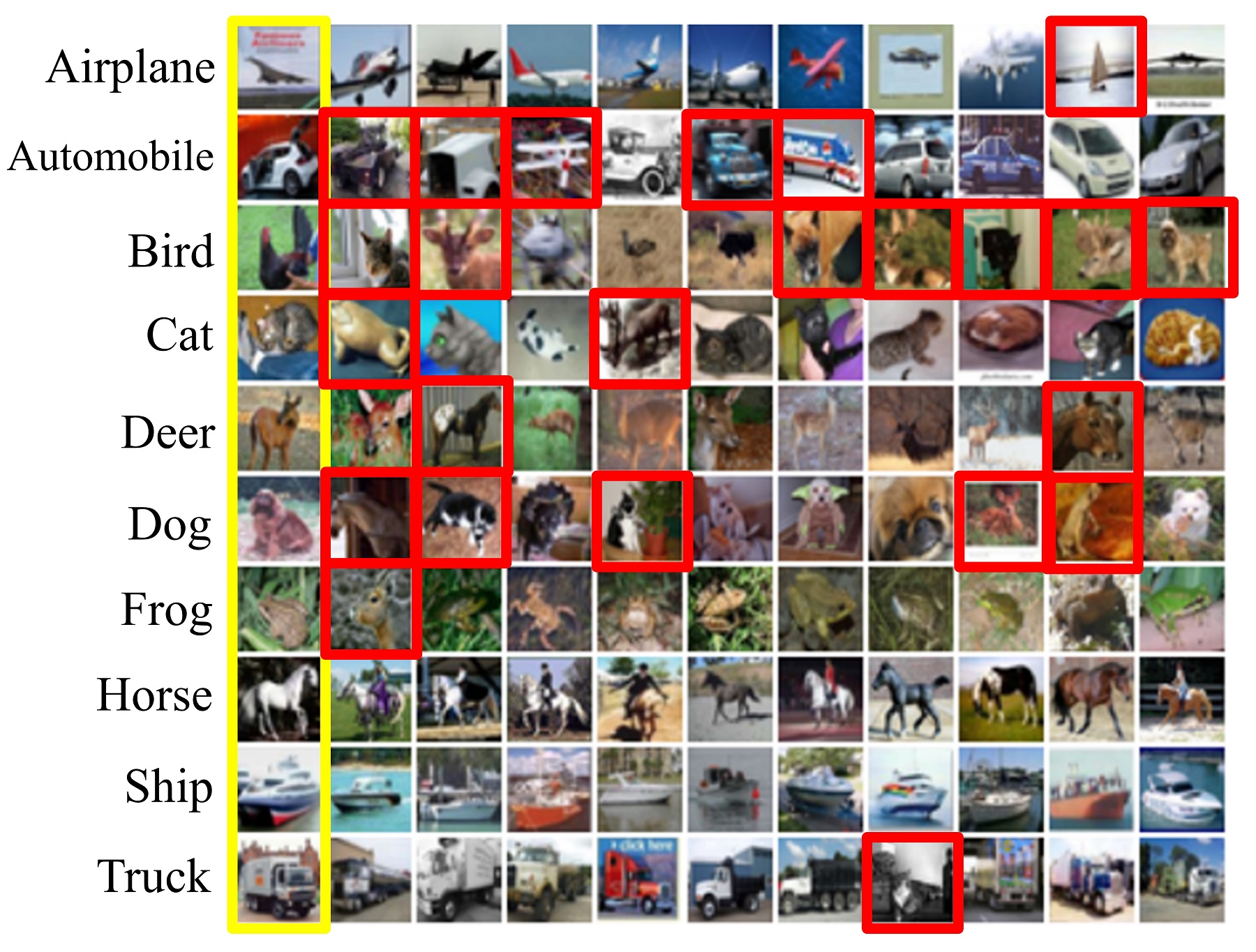}
			\label{search_cifar_npair}}
		\caption{Image search application on CIFAR-10: The first column (yellow box) represents the query image, the remaining columns are the 10 nearest images of the query image, and the irrelevant image result is marked by the red box.}
		\label{fig:image_search_cifar}
	\end{figure*}

\begin{figure*}[t]
		\centering
		\subfloat[CovNet v1]{\includegraphics[width=0.28\linewidth]{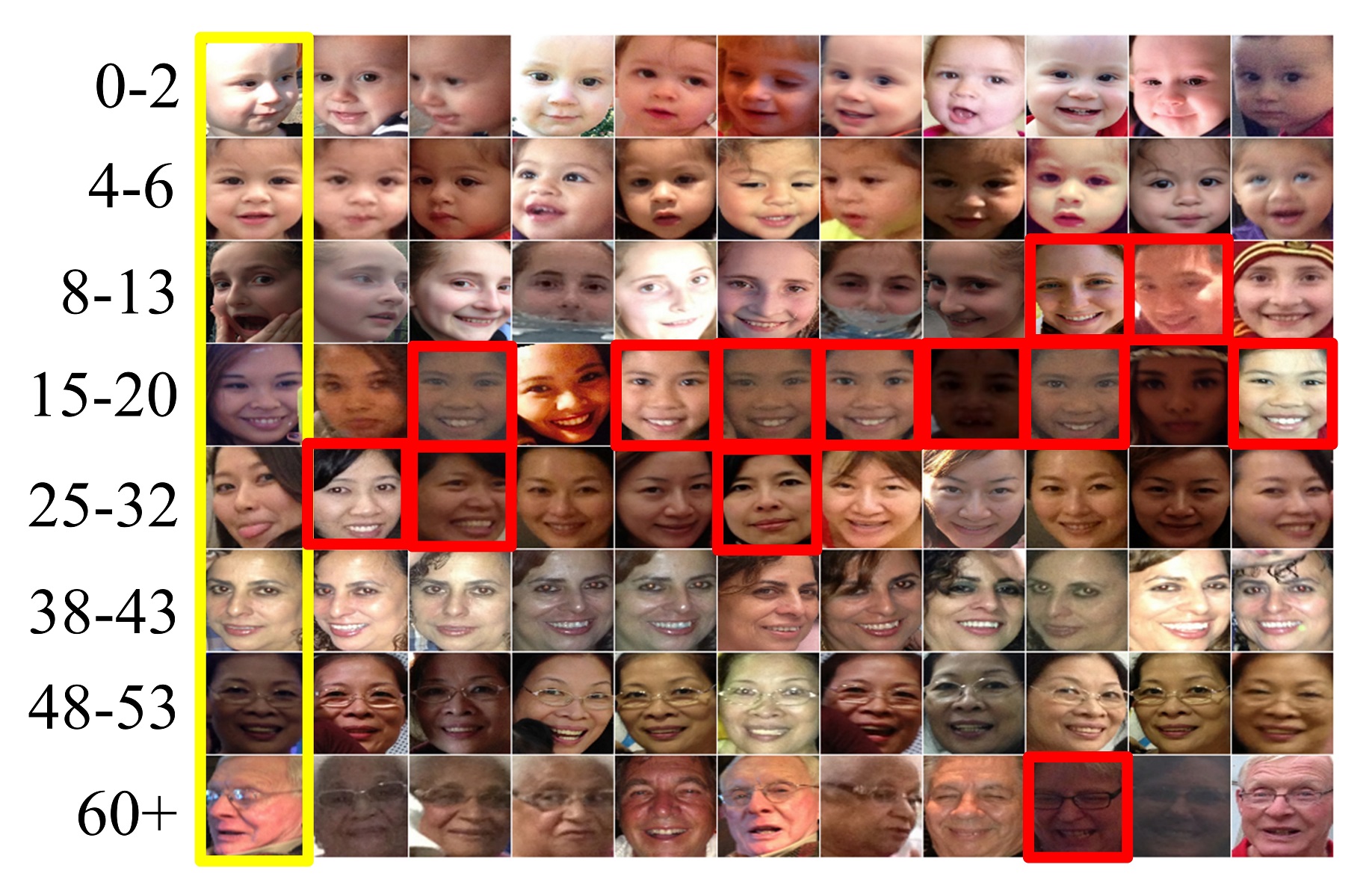}
			\label{search_adience_v1}}
		\subfloat[CovNet v2]{\includegraphics[width=0.28\linewidth]{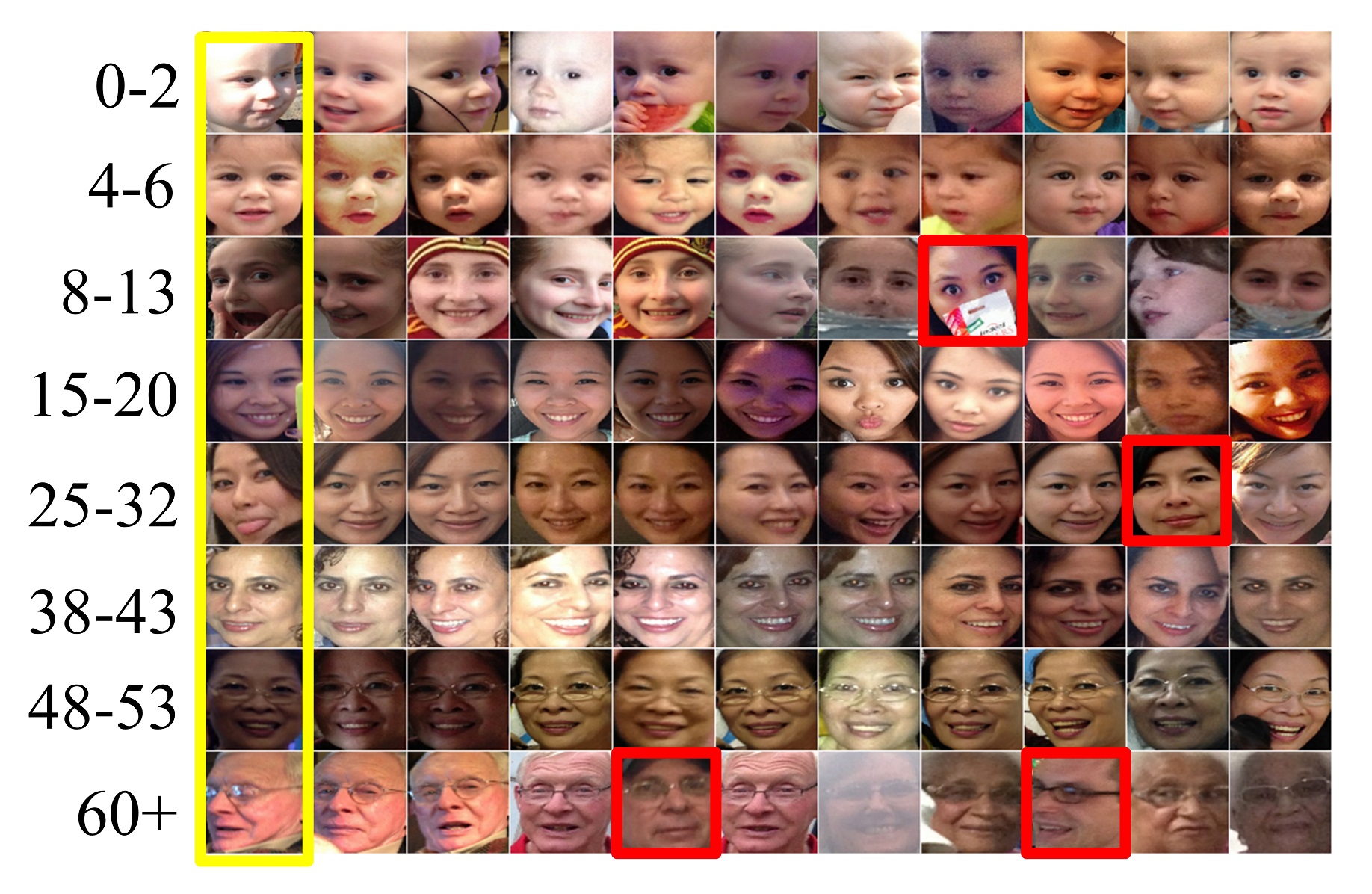}
			\label{search_adience_v2}}
		\subfloat[CovNet v3]{\includegraphics[width=0.28\linewidth]{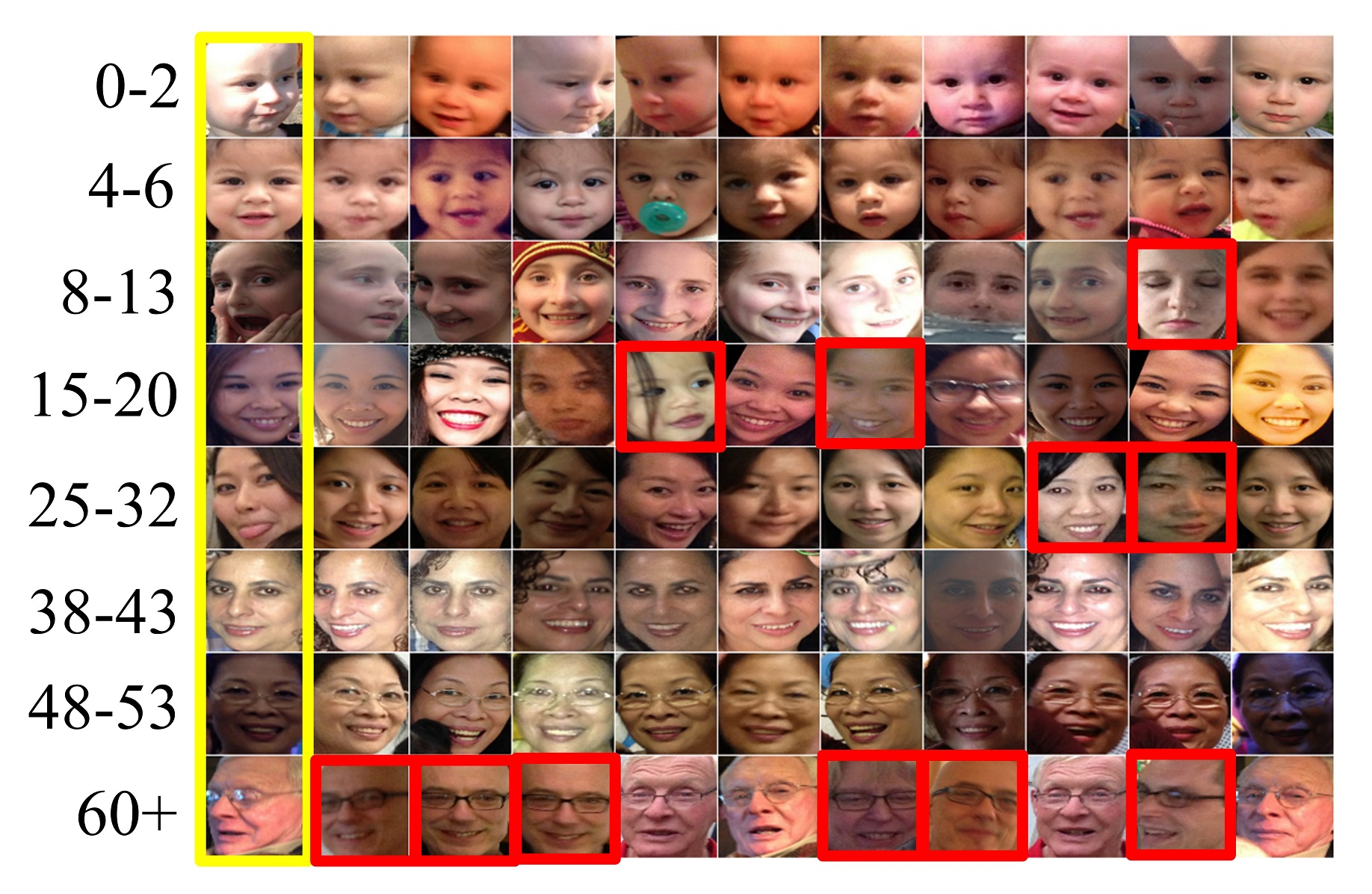}
			\label{search_adience_v3}}
		
		\subfloat[Siamese]{\includegraphics[width=0.28\linewidth]{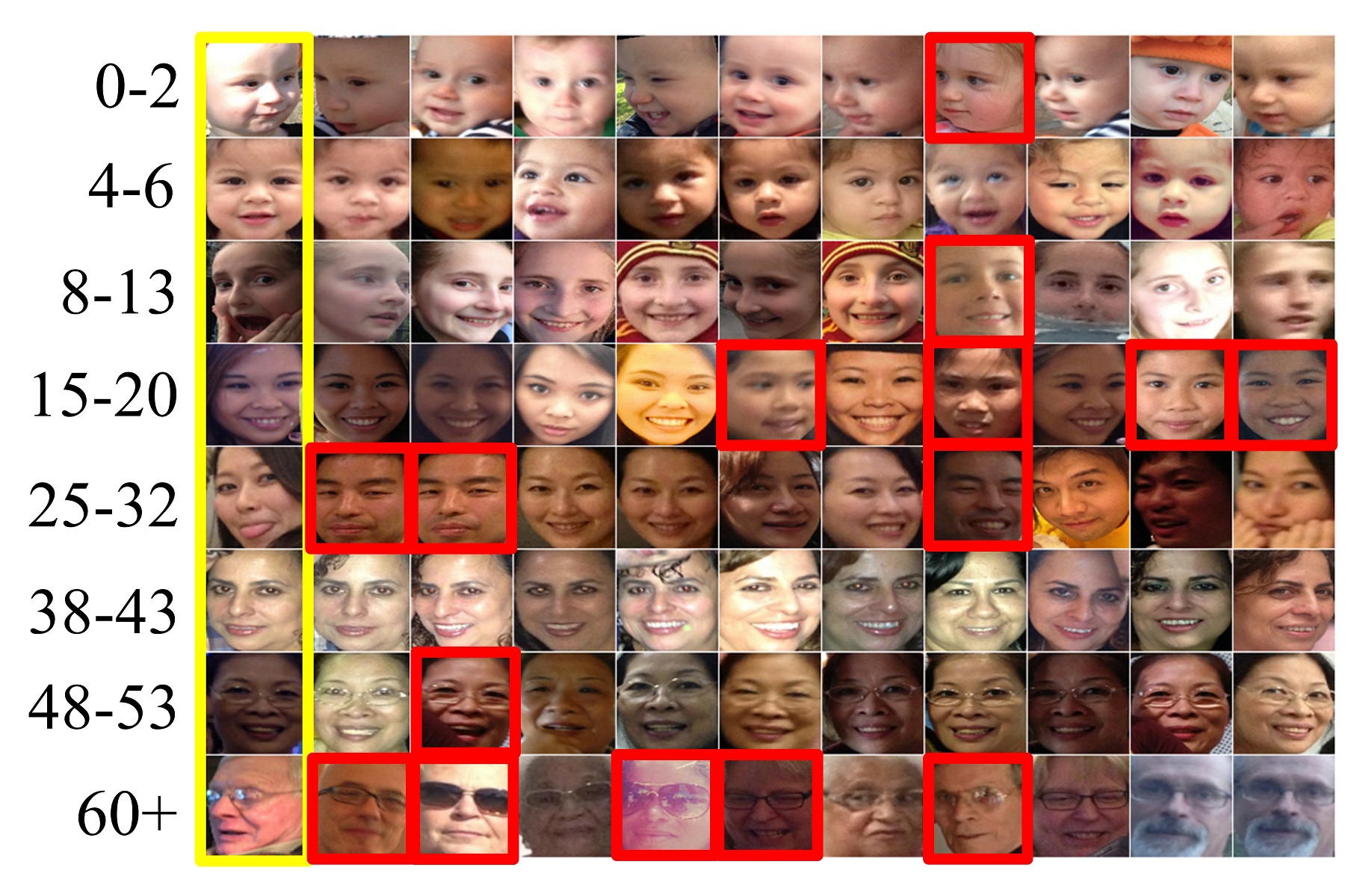}
			\label{search_adience_siamese}}
		\subfloat[Triplet]{\includegraphics[width=0.28\linewidth]{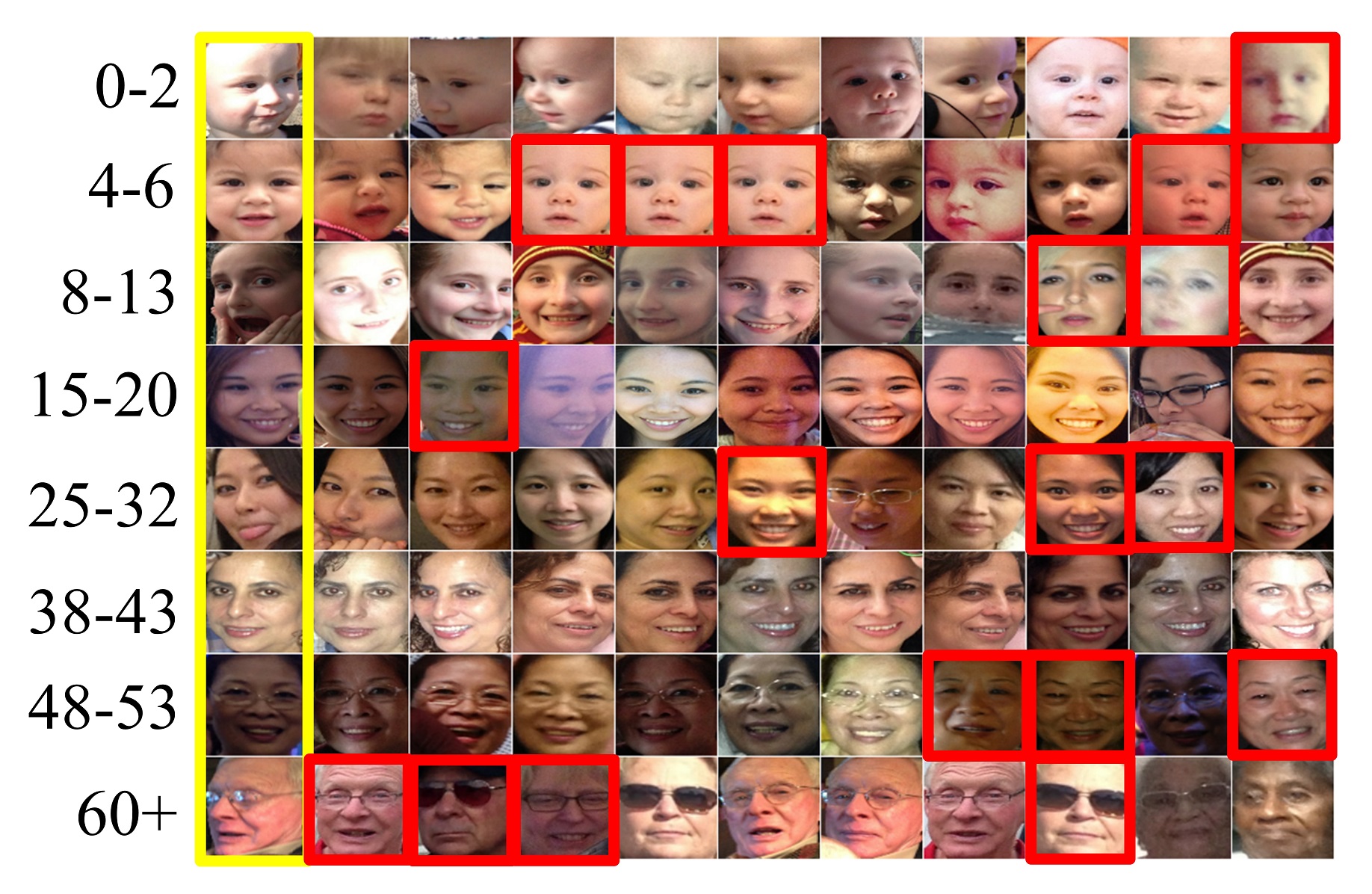}
			\label{search_adience_triplet}}
		\subfloat[N-Pair]{\includegraphics[width=0.28\linewidth]{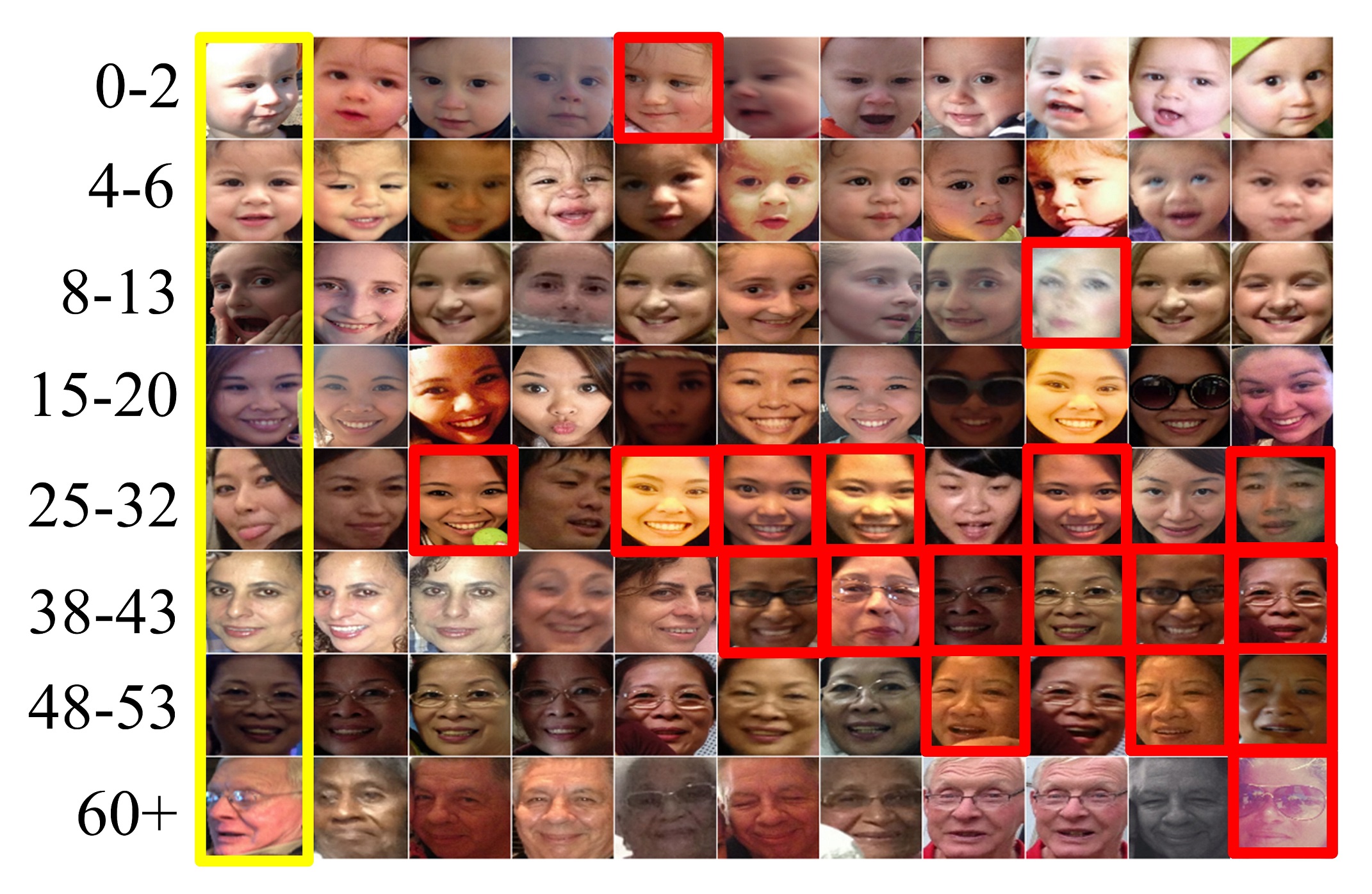}
			\label{search_adience_npair}}
		\caption{Image search application on Adience: The first column (yellow box) represents the query image, the remaining columns are the 10 nearest images of the query image, and the irrelevant image result is marked by the red box.}
		\label{fig:image_search_adience}
	\end{figure*}

	As a service, the goal of computing technology is to perform business services more efficiently and effectively. Nowadays, many real-world applications must deal with large-scale datasets, such as image similarity searches and content-based recommendation systems. Thus, metric learning can be efficient because it can project high-dimensional images into a small vector. A vector representation must also reflect the original data effectively. One of the interesting applications of near-neighbor (or similarity) problems is image search similarity. In this case, given a query image, the model retrieves $k$ similar images as the query image. The visualization of the image similarity is illustrated in Figs. \ref{fig:image_search_cifar} and \ref{fig:image_search_adience}. The first column (yellow box) represents the query image, whereas the remaining columns correspond to the top-10 similar images. The red box represents irrelevant images based on the label of the query image. In Fig. \ref{fig:image_search_cifar},  the irrelevant images mainly occur on dogs-cats and deer-horses, which share similar features. In Fig. \ref{fig:image_search_adience}, because the label (face age) naturally has an ordinal relationship, the error commonly occurs in the nearest age; for instance, 15–20 year-olds are often predicted as 25–32 year-olds. In general, we can infer that CovNet v2 outperforms the other models. Here, CovNet v1 and v3 slightly outperformed the Siamese network. Compared with the triplet and N-pair networks, the Siamese network is more competitive with our proposed models.
	
	\subsection{Semantic analysis}
	
\begin{figure*}[!t]
		\centering
		\subfloat[CovNet v1]{\includegraphics[width=0.25\linewidth]{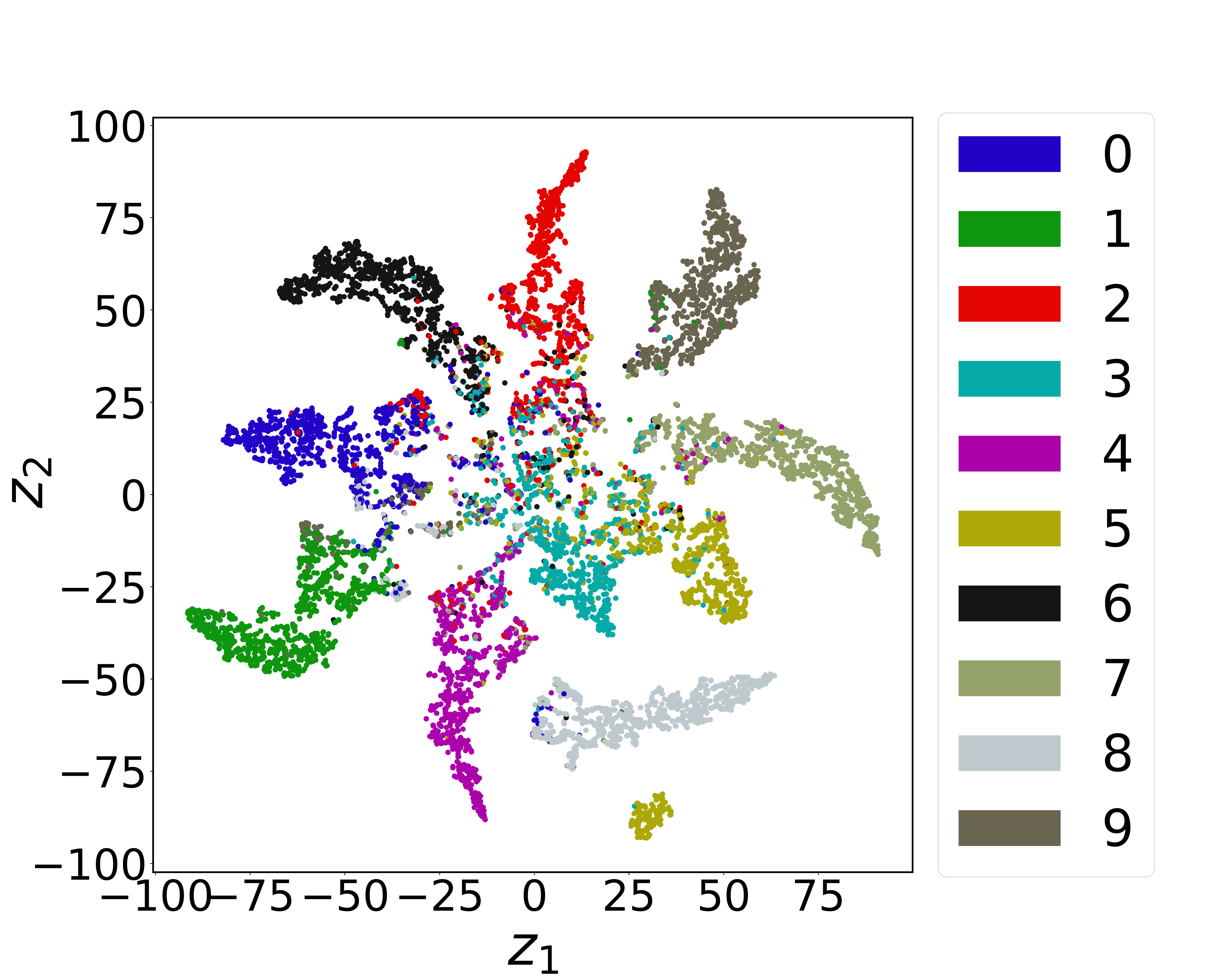}
			\label{latent_cifar_v1}}
		\subfloat[CovNet v2]{\includegraphics[width=0.25\linewidth]{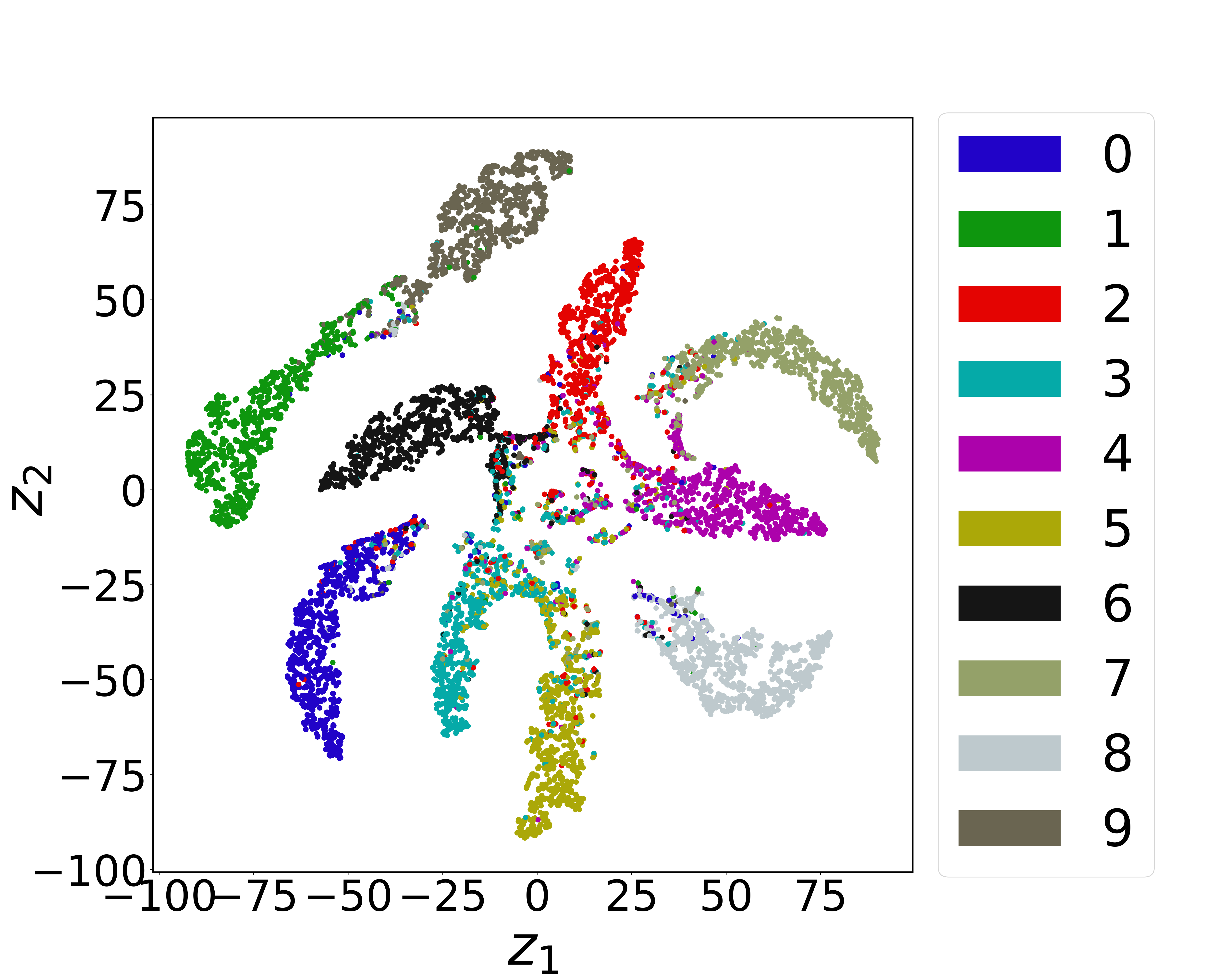}
			\label{latent_cifar_v2}}
		\subfloat[CovNet v3]{\includegraphics[width=0.25\linewidth]{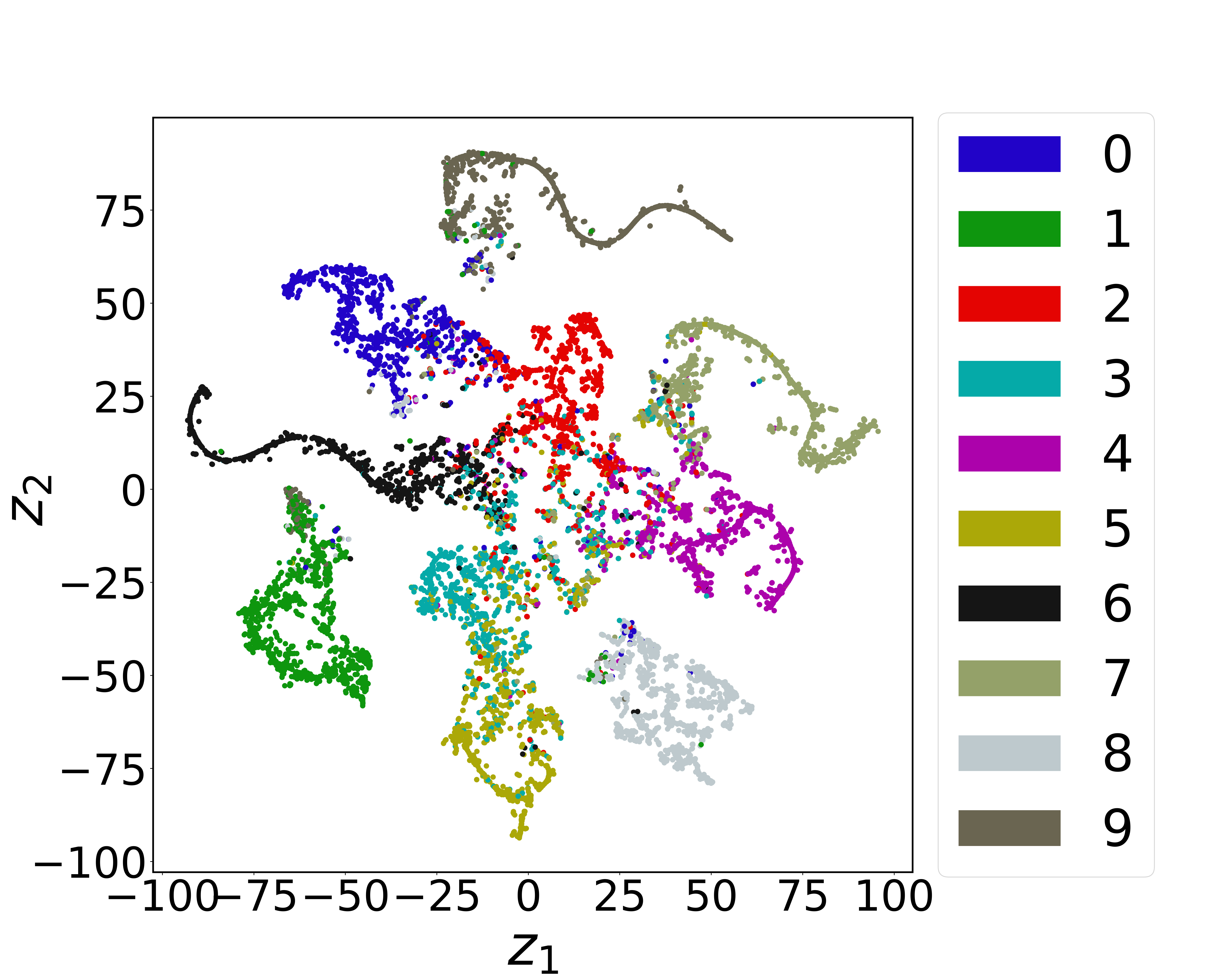}
			\label{latent_cifar_v3}}
		
		\subfloat[Siamese]{\includegraphics[width=0.25\linewidth]{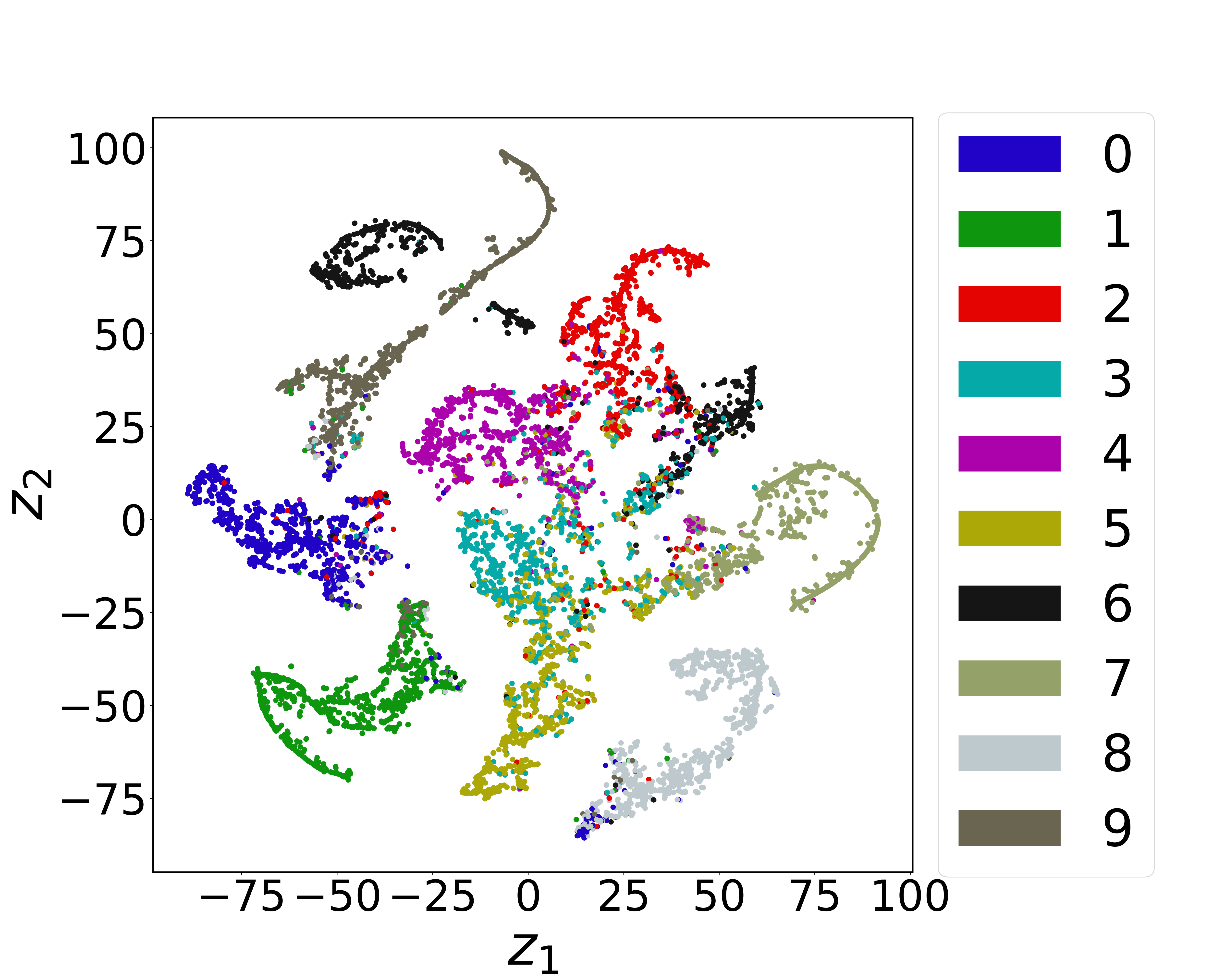}
			\label{latent_cifar_siamese}}
		\subfloat[Triplet]{\includegraphics[width=0.25\linewidth]{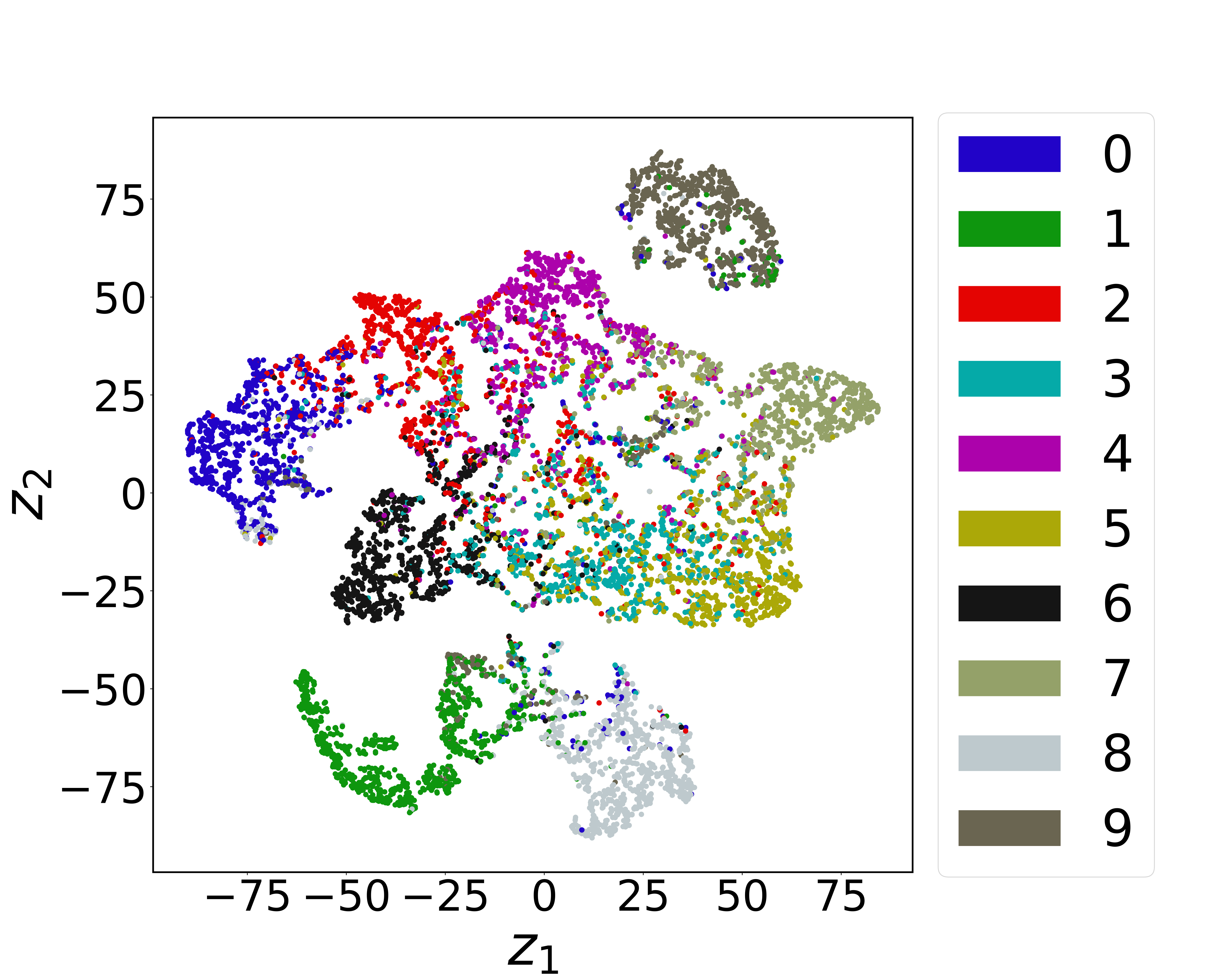}
			\label{latent_cifar_triplet}}
		\subfloat[N-pair]{\includegraphics[width=0.25\linewidth]{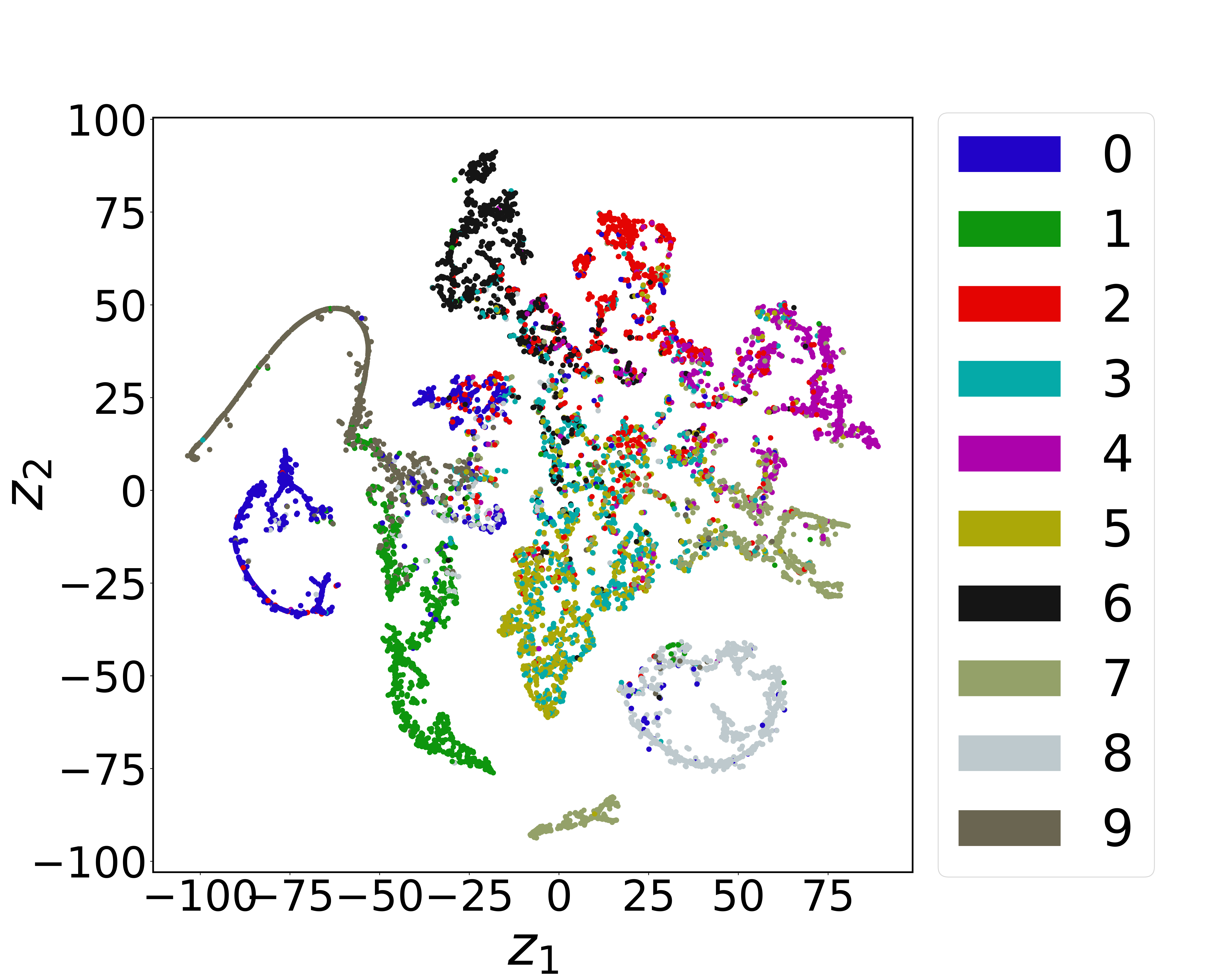}
			\label{latent_cifar_npair}}
		\caption{Visualization of latent separability on CIFAR-10 by projecting an embedding space to a two-dimensional space using t-SNE.}
		\label{fig:latent_vis_cifar}
	\end{figure*}

	In this section, the semantic relationships between both inter-class separability and inter-class relationships are observed. We present Fig. \ref{fig:latent_vis_cifar} as a visualization of the embedding space to assess inter-class separability. The output of the embedding network was a vector with 100 dimensions. Accordingly, we utilized t-stochastic embedding (t-SNE) to project them into a two-dimensional space. As illustrated in Fig. \ref{fig:latent_vis_cifar}, our proposed methods have a better embedding space separability compared with Siamese, Triplet, and N-pair networks. Therefore, we can reveal that covariance embedding generates separable outcomes more effectively than the Euclidean and cosine similarity metrics. Note that, unlike the Euclidean and cosine similarity metrics, covariance embedding is more expressive in capturing the relationship between two inputs because it can reflect the positive, negative, or even neutral correlation simultaneously. In addition, for the merging layer (see Table \ref{tab:total_model_comp}), CovNet obtains a vector that is directly assessed using categorical or binary cross-entropy. By contrast, other models result in a scalar in the merging layer. Thus, this mechanism can be more effective in preserving the data separability in the embedding space because assessing a vector (multivariate) as a feature is more effective for guaranteeing separability than representing it as a scalar (univariate).
		
	\begin{figure*}[!h]
		\centering
		\subfloat[CovNet v1]{\includegraphics[width=0.25\linewidth]{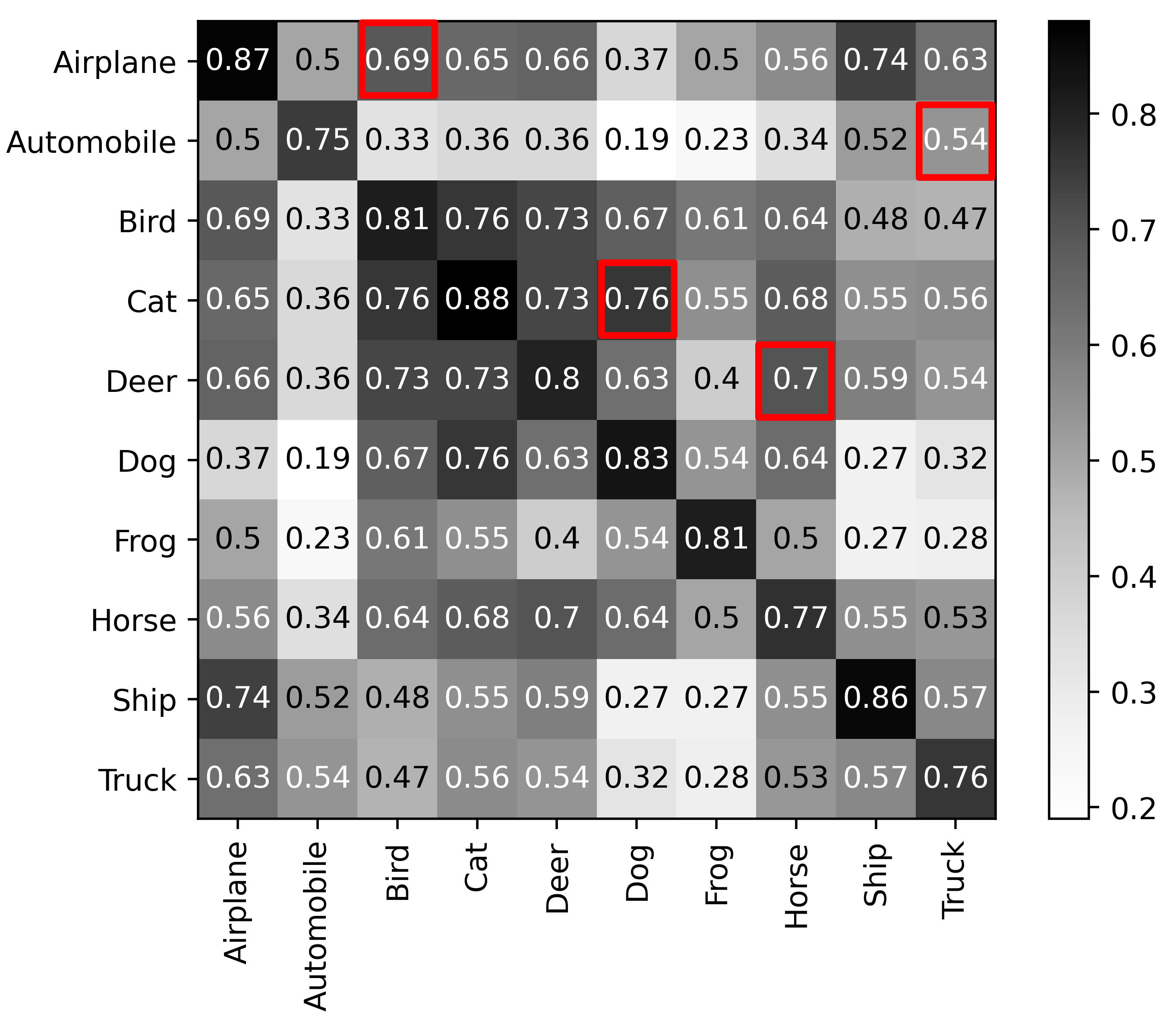}
			\label{corr_covnet_v1}}
		\subfloat[CovNet v2]{\includegraphics[width=0.25\linewidth]{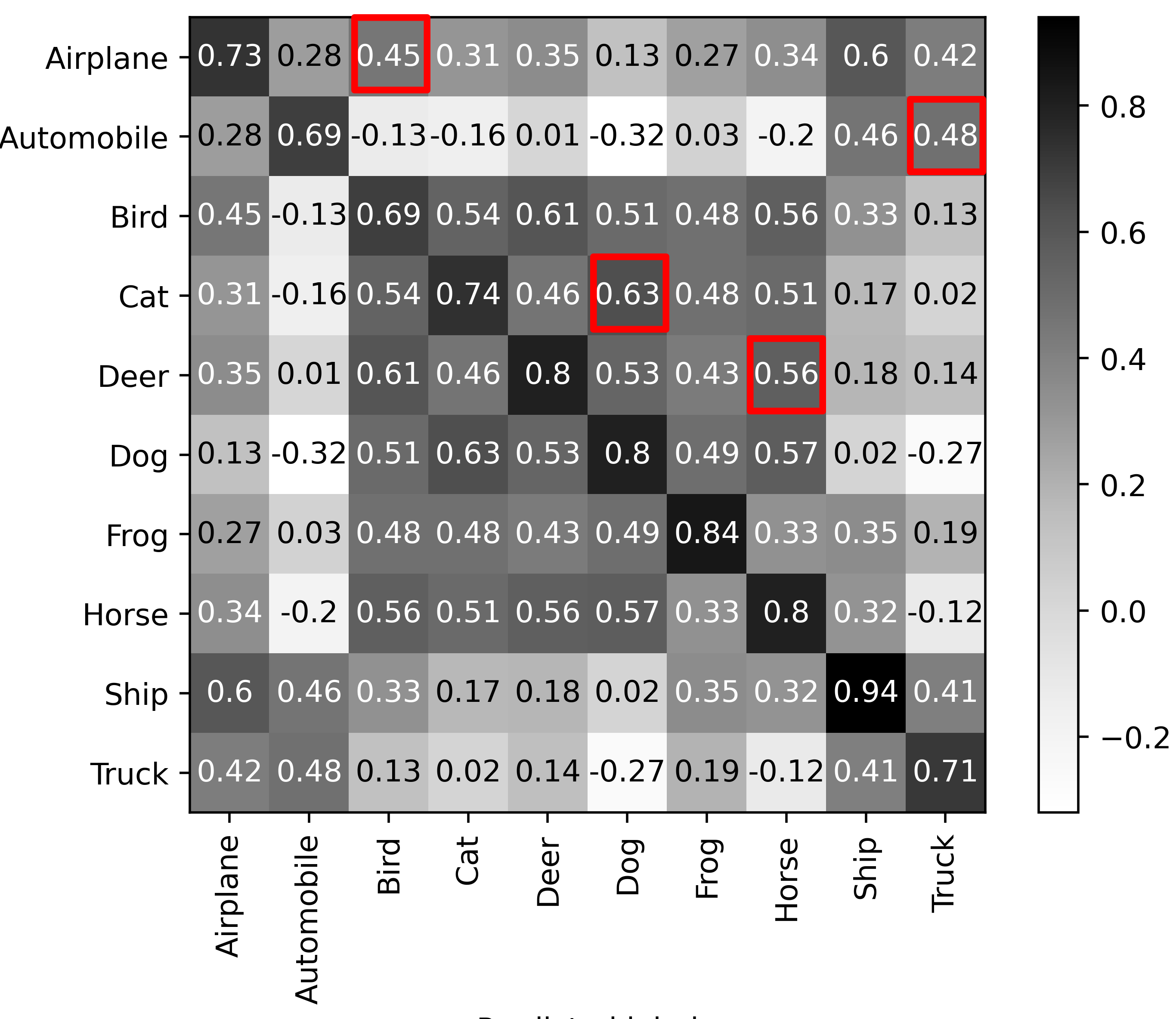}
			\label{corr_covnet_v2}}
		\subfloat[CovNet v3]{\includegraphics[width=0.25\linewidth]{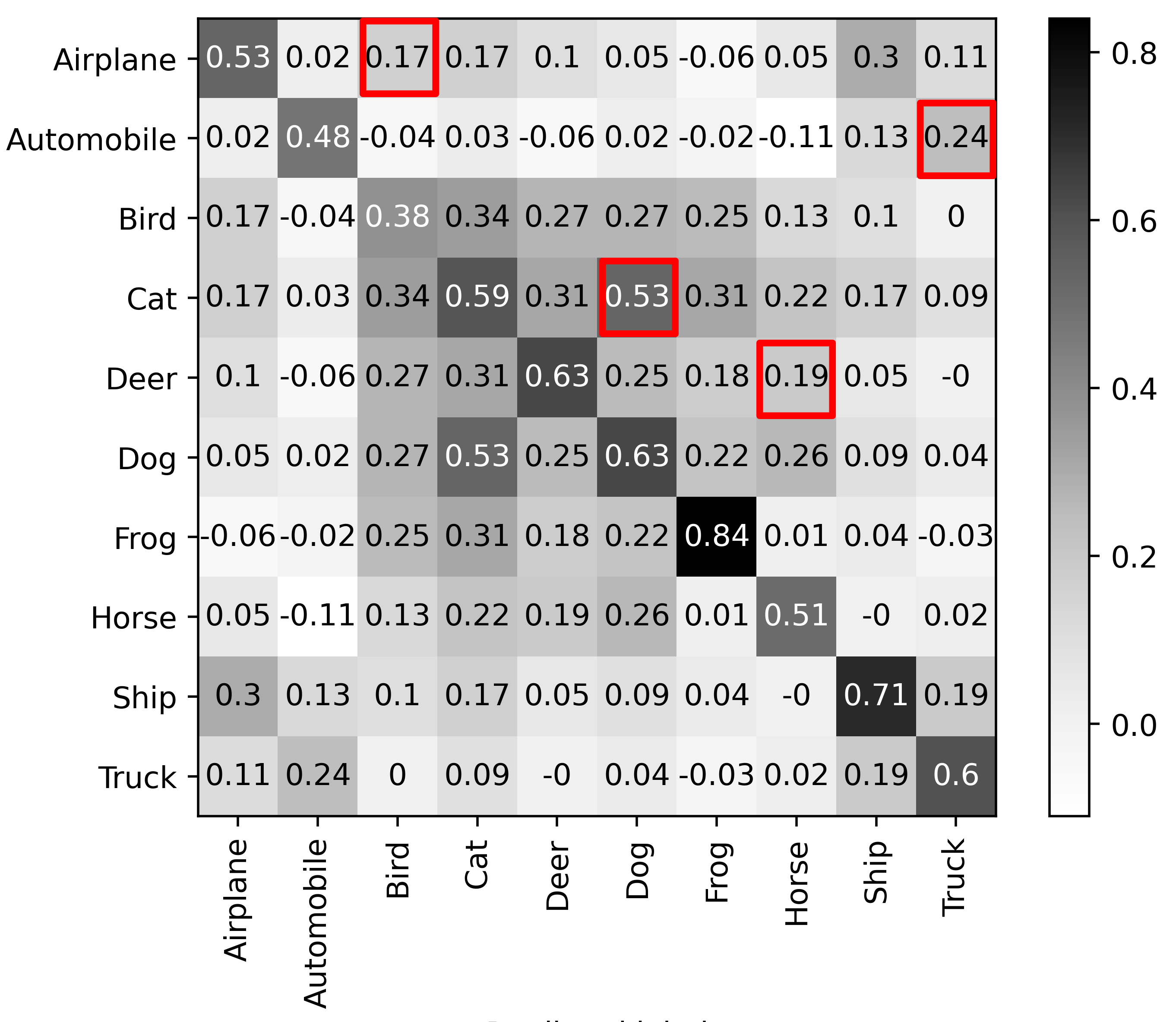}
			\label{corr_covnet_v3}}
		
		\subfloat[Siamese]{\includegraphics[width=0.25\linewidth]{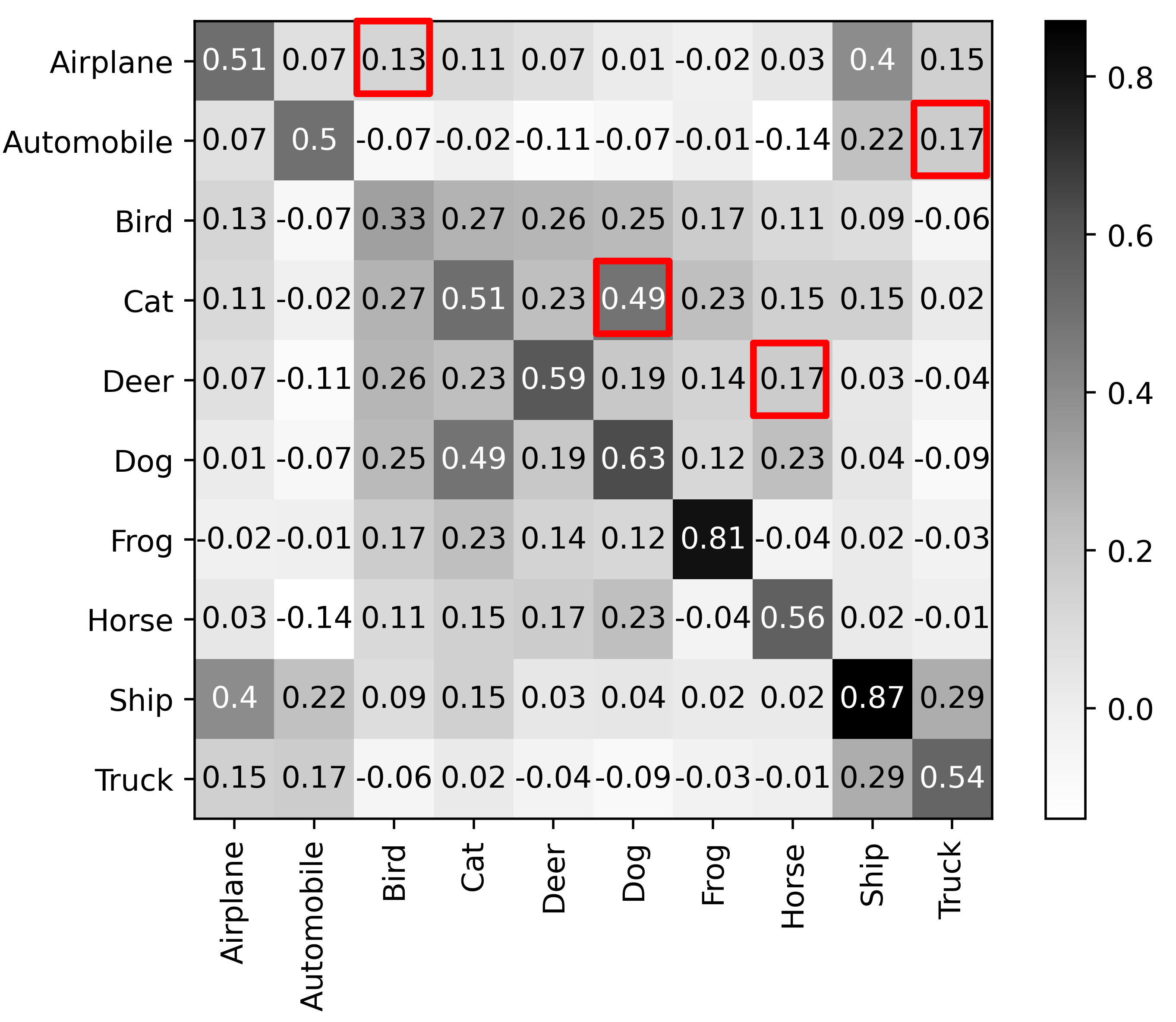}
			\label{corr_siamese}}
		\subfloat[Triplet]{\includegraphics[width=0.25\linewidth]{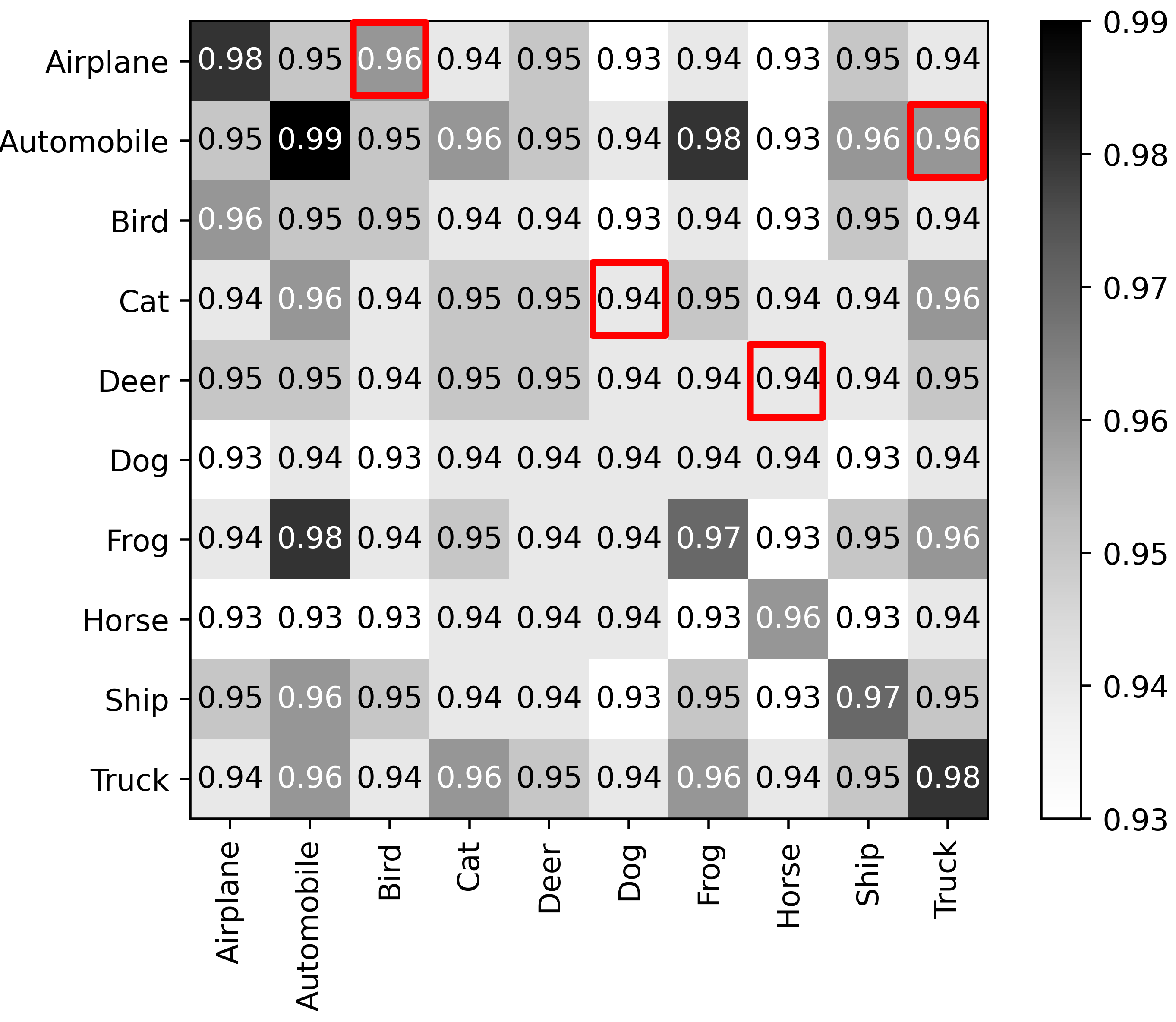}
			\label{corr_triplet}}
		\subfloat[N-pair]{\includegraphics[width=0.25\linewidth]{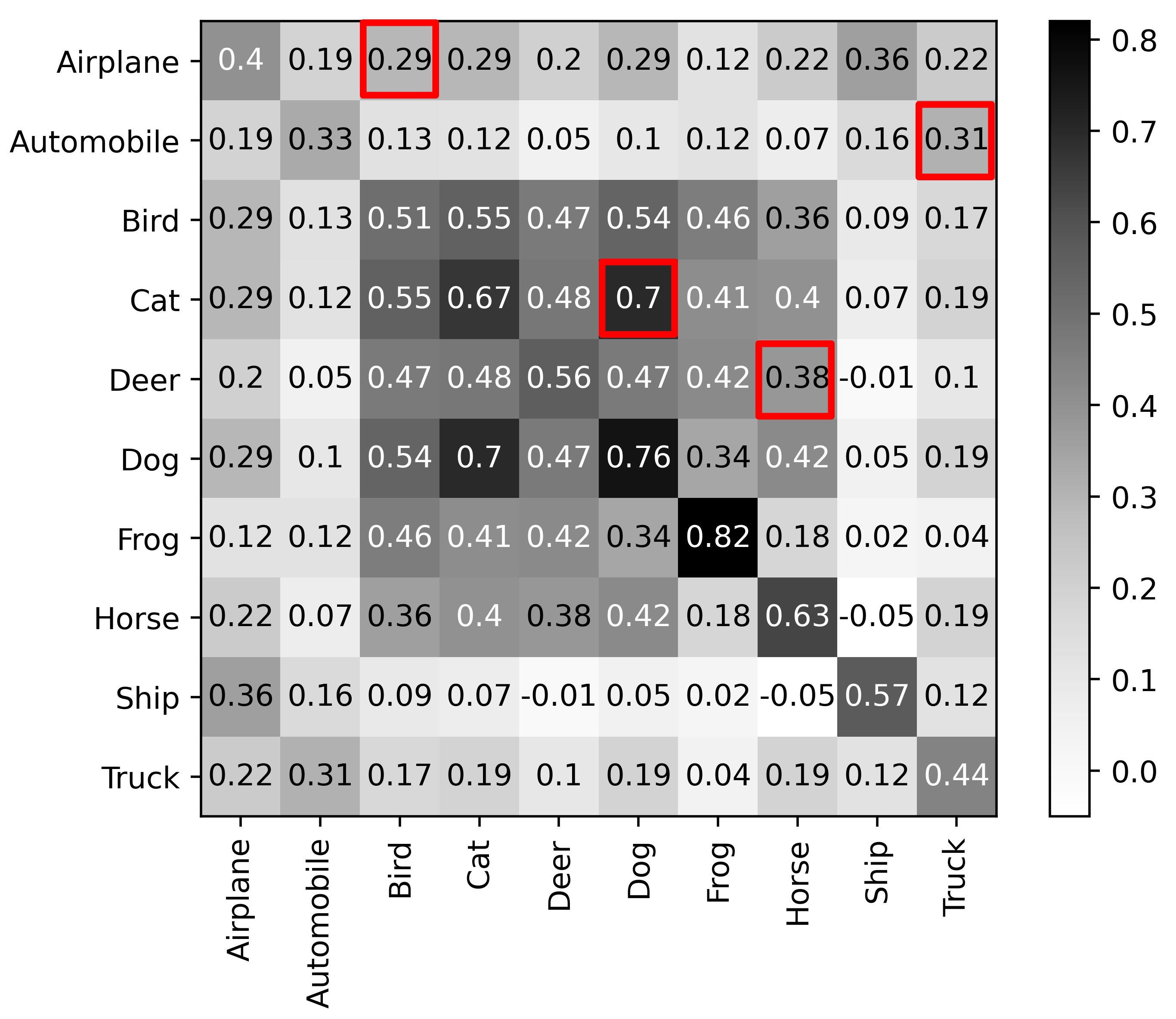}
			\label{corr_npair}}
		\caption{Pearson correlation matrix to capture the semantic relationship among categories on CIFAR-10.}
		\label{fig:corr_matrix}
	\end{figure*}

	In addition, we argue that a reliable embedding-space representation can be assessed not only through inter-class separability but also by inter-class relationships. For instance, semantically, we agree that dog-cat and truck-automobile share similar shapes and properties. Therefore, a good embedding-space representation must also preserve this special proximity property. This property is useful for content-based recommendation systems and search engine applications. For better services, in addition to recommending similar products, it is better if the system also offers different products that have similar characteristics as the user query. Consequently, we employed the Pearson correlation method to assess the relationships among the inter-class images. The results are summarized in a correlation matrix, as shown in Fig. \ref{fig:corr_matrix}. Compared with other models, our CovNet is more expressive in describing an inter-class relationship. Unlike the existing models, the inter-class correlation value is still high, whereas the value is still lower than the inner-class correlation value. Inherently, a well-known similar object in CIFAR-10 is in a different class, such as cat-dog, airplane-bird, deer-horse, and automobile-truck (red box in Fig. \ref{fig:corr_matrix}). Nevertheless, unlike our CovNet (particularly v1 and v2), they have low correlation values in Siamese, Triplet, and N-pair networks. The correlation matrix developed using Siamese is slightly similar to that of CovNet v3 because they have a similar architecture. However, semantically, CovNet v3 has a slightly higher correlation in terms of inter-class relationships than Siamese. In the triplet network, the inter-class correlation value is too aggressive, and it has a higher magnitude than its inner-class correlation. Moreover, the N-pair network is too fierce in defining inter-class relationships, and thus some inner-class correlation values also become low.
	
	\begin{table*}[!t]
		\centering
		\caption{Inspection of embedding space separability using a standard classifier (support vector machine).}
		\label{tab:classification_svm}
		\resizebox{0.67\linewidth}{!}{
			\centering
			\begin{tabular}{l l l l l l}		
				\toprule
				\multirow{2}{*}{Model} & \multicolumn{5}{c}{Accuracy \% (mean $\pm$ standard deviation)}\\
				\cmidrule{2-6}
				&	CIFAR-10 			&	Food-11				&	Colorectal			&	Adience				&	Yale \\
				\midrule
				CovNet v1			&	$85.4\pm0.48$		&	$81.2\pm0.72$		&	$84.7\pm0.53$		&	$65.3\pm0.49$		&	$100.0\pm0$	 \\
				CovNet v2			&	$86.1\pm0.39$		&	$82.3\pm0.65$		&	$86.5\pm0.46$		&	$66.0\pm0.37$		&	$100.0\pm0$  \\
				CovNet v3			&	$85.2\pm0.41$		&	$79.9\pm0.70$		&	$83.6\pm0.67$		&	$64.4\pm0.43$		&	$100.0\pm0$	 \\				
				Siamese				&	$85.2\pm0.42$		&	$79.0\pm0.69$		&	$82.1\pm0.50$		&	$64.4\pm0.47$		&	$100.0\pm0$	 \\
				Triplet				&	$74.1\pm0.61$		&	$36.6\pm0.94$		&	$77.0\pm0.73$		&	$65.0\pm0.68$		&	$100.0\pm0$	 \\
				N-pair				&	$71.7\pm0.58$		&	$48.9\pm0.74$		&	$48.5\pm0.65$		&	$62.1\pm0.65$		&	$100.0\pm0$  \\
				N-pair(1000)		&	$83.5\pm0.35$		&	$75.1\pm0.67$		&	$67.2\pm0.59$		&	$64.6\pm0.37$		&	$100.0\pm0$	 \\
				\bottomrule
			\end{tabular}
		}
	\end{table*}
	
	\subsection{Classification and parameter}
	Because of their effectiveness, several metric learning models, such as Siamese, Triplet, and N-pair networks, are commonly used for fine-tuning classification. In this study, a metric learning model acts as a feature extraction network (base model). Here, we train only a few layers on top of it. Meanwhile, the weights of the pre-trained network were not updated during the training. Commonly, we use a deep neural network (DNN) as the top model to classify the data. However, because of its stochastic nature, and for the sake of a fair comparison, we employ a standard classifier, i.e., an SVM with a radial basis function (RBF) kernel, to classify the embedding outcome. The classification performances are presented in Table \ref{tab:classification_svm}. In line with the previous analysis results in the previous sections, CovNet v2 has the best performance compared with other metric learning models. Siamese and CovNet v3 have a competitive performance because their architectures are quite similar. The triplet network is slightly better than N-Pair, except in Food-11. In addition, increasing the number of batches in the N-pair network can increase the classification performance, except in the Yale dataset, which is a relatively small dataset (165 instances). Thus, it was unaffected by the number of batches. Moreover, all models become competitive with each other if we utilize a pre-trained network, as shown in the Adience and Yale datasets.
	
	\begin{table}[!t]
		\centering
		\caption{Comparison of the number of trained parameters in each model.}
		\label{tab:total_parameter}
		\resizebox{1\linewidth}{!}{
			\begin{threeparttable}
				\centering
				\begin{tabular}{l l l l l l}		
					\toprule
					\multirow{2}{*}{Model} & \multicolumn{5}{c}{Number of trainable parameters}\\
					\cmidrule{2-6}
					&	CIFAR-10 			&	Food-11				&	Colorectal			&	Adience\tnote{*}	&	Yale\tnote{*} 	\\
					\midrule
					CovNet v1			&	1,830,294			&	2,486,523			&	2,486,523			&	119,180			&	119,887		\\
					CovNet v2			&	1,834,839			&	2,492,078			&	2,489,078			&	122,008			&	130,492		\\
					CovNet v3			&	1,829,585			&	2,485,713			&	2,485,713			&	118,673			&	118,673		\\
					Siamese				&	1,829,585			&	2,485,713			&	2,485,713			&	118,673			&	118,673		\\
					Triplet				&	1,829,284			&	2,485,412			&	2,485,412			&	970,340			&	128,472		\\
					N-pair				&	1,829,284			&	2,485,412			&	2,485,412			&	970,340			&	128,472		\\
					\bottomrule
				\end{tabular}
				\begin{tablenotes}
					\item[*] Employing FaceNet \cite{7298682} (pre-trained network) as a base model.
				\end{tablenotes}
			\end{threeparttable}
		}
	\end{table}
	
	Finally, we present a comparison of the number of trainable parameters in each model, as described in Table \ref{tab:total_model_comp}. In general, the number of parameters in each model is not significantly different. CovNet v2 has a higher trained parameter because the tail layer has more neuron units compared with the others, as described in Section \ref{sec:experiment_setting}. CovNet v3 and Siamese have the same numbers of parameters because they have the same architecture except for the merging layer. The merging layer in Table \ref{tab:total_model_comp} has no parameters. The triplet and N-pair networks have the least number of parameters because the triplet loss and cosine similarity loss can be directly optimized such that a tail layer is not required. Nevertheless, their performance is significantly lower than that of the other models. The numbers of parameters in the Adience and Yale datasets are less than the others because they employ a pre-trained network, FaceNet, as a base model.
	
	\section{Conclusion}
	\label{sec:conclusion}
	A desirable metric learning model should not only separate the dataset based on its categories it should also maintain the inter-class semantic relationship. We present CovNets, a novel metric learning method, as a service approach with covariance embedding optimization. Through the development of CovNet, we show two important properties: (1) A desirable metric learning model should not only separate the dataset based on its categories but also maintain the inter-class semantic relationship. (2) Covariance-enhanced embedding can make complex machine learning tasks more expressive, more explainable, and more efficient. For example, by producing covariance embedding, face verification simply involves thresholding the distance between two embeddings. Similarly, object recognition, recommender systems, and an image search similarity when producing covariance embedding with CovNets can be reduced to a $k$-NN classification problem. High-dimensional data clustering with CovNets feature-engineering through covariance embedding can be reduced to a linear-space problem solvable using simple bottom-up techniques, including agglomerative clustering. Extensive empirical experiments demonstrate the effectiveness of CovNets over representative state-of-art metric learning models, including Siamese, Triplet, and N-pair networks. 
	
	\section*{Acknowledgment}
	This research was supported by the National Research Foundation of Korea(NRF) grant funded by the Korea government(MSIT) (No. 2020R1A2C1102294).

	\ifCLASSOPTIONcaptionsoff
	\newpage
	\fi

	
	
	\bibliographystyle{IEEEtran}
	\bibliography{bare_jrnl_compsoc}
	%
	
	
	
	%

	
	

\end{document}